\documentclass[runningheads]{llncs}

\usepackage{float}
\usepackage{caption}
\usepackage{subcaption}
\usepackage{multirow}
\usepackage{array}
\usepackage{eccv}

\usepackage{eccvabbrv}

\usepackage{graphicx}
\usepackage{booktabs}
\usepackage{soul}

\usepackage[accsupp]{axessibility}  % Improves PDF readability for those with disabilities.

\usepackage{orcidlink}

\usepackage[inkscapelatex=false]{svg}

\begin{document}

\title{StyleSplat: 3D Object Style Transfer with Gaussian Splatting}
% \title{Dynamic Gaussian Styling: 3D Object Style Transfer of Dynamic Gaussian}

% \titlerunning{StyleSplat: 3D Object Style Transfer with Gaussian Splatting}

\author{Sahil Jain\thanks{equal contribution}  \and
Avik Kuthiala$^\star$ \and
Prabhdeep Singh Sethi\and
Prakanshul Saxena}

\authorrunning{S. Jain et al.}

\institute{Carnegie Mellon University}

\maketitle
\begin{abstract}
Recent advancements in radiance fields have opened new avenues for creating high-quality 3D assets and scenes. Style transfer can enhance these 3D assets with diverse artistic styles, transforming creative expression. However, existing techniques are often slow or unable to localize style transfer to specific objects. We introduce StyleSplat, a lightweight method for stylizing 3D objects in scenes represented by 3D Gaussians from reference style images. Our approach first learns a photorealistic representation of the scene using 3D Gaussian splatting while jointly segmenting individual 3D objects. We then use a nearest-neighbor feature matching loss to finetune the Gaussians of the selected objects, aligning their spherical harmonic coefficients with the style image to ensure consistency and visual appeal. StyleSplat allows for quick, customizable style transfer and localized stylization of multiple objects within a scene, each with a different style. We demonstrate its effectiveness across various 3D scenes and styles, showcasing enhanced control and customization in 3D creation.

\end{abstract}

\section{Introduction}
Breakthroughs in radiance field generation have revolutionized the capture and representation of 3D scenes, allowing for unprecedented levels of detail and realism. The ability to seamlessly transfer artistic styles to objects extracted from these radiance fields offers a transformative approach for industries such as gaming, virtual reality, and digital art. This technology not only enhances creative expression but also significantly reduces the time and effort required to produce visually stunning 3D assets, pushing the boundaries of what is possible in digital content creation.

The emergence of 3D Gaussian splatting (3DGS) \cite{Kerbl20233DGS} has introduced a powerful method for representing radiance fields, offering the advantage of fast training and rendering while preserving high-quality details. Prior to this advancement, neural radiance field (NeRF) based methods \cite{mipnerf, mildenhall2020nerfrepresentingscenesneural, mueller2022instant} have been extensively utilized for reconstructing detailed scenes, and many techniques \cite{lahiri2023s2rf, li2023arfpluscontrollingperceptualfactors, coarf} aim at transferring artistic styles to individual objects within NeRFs. However, the slow training and rendering speeds of NeRFs pose significant challenges for their practical use.

\label{sec:intro}
\begin{figure}[!htbp]
    \centering
    \includegraphics[width=\linewidth]{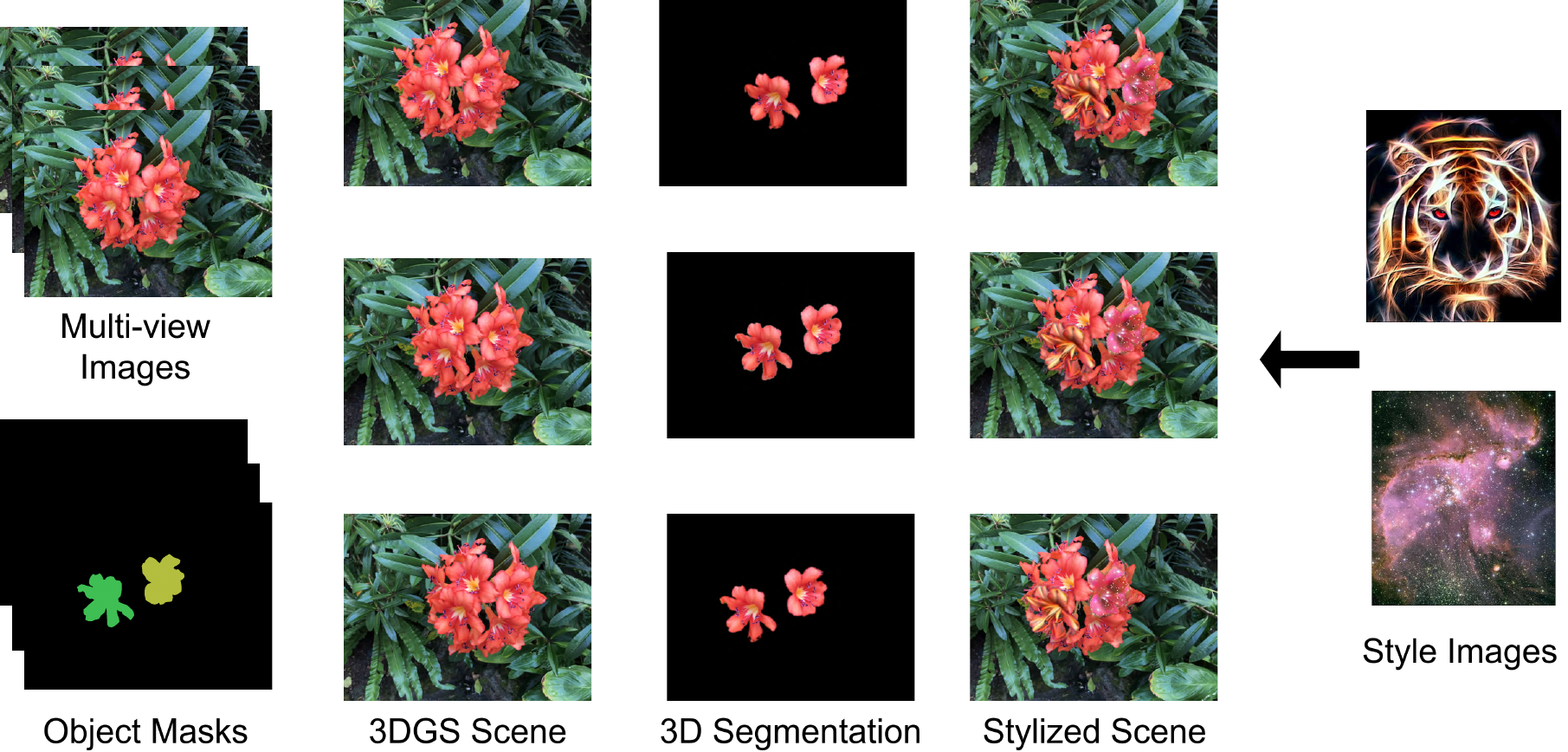}
    \caption{We introduce StyleSplat, an approach for lightweight, customizable, and localized stylization of 3D objects from reference style images. Our approach first learns a photorealistic representation of the scene with 3D Gaussian splatting while jointly segmenting the scene into individual 3D objects using 2D masks. We then employ a nearest-neighbor feature matching loss to finetune and stylize the user-specified objects using the provided style images.}
\end{figure}

Approaches have also been introduced in the 3DGS paradigm \cite{liu2024stylegaussian,gsinstyle} that enable the transfer of style in real time. However, they apply the style globally to entire scenes without providing mechanisms for localized application to individual objects. Furthermore, although text-based approaches \cite{igs2gs} can facilitate object-specific style transfer through text instructions using an image-conditioned diffusion model, the reliance on textual descriptions introduces ambiguities, particularly in accurately conveying specific colors, styles, or textures within 3D scenes.

In this paper, we introduce StyleSplat, a lightweight method for stylizing 3D objects in scenes represented using 3D Gaussians from arbitrary style images. Our approach allows for the stylization of multiple objects within a scene, each with a different style image. StyleSplat consists of three steps: 2D mask generation and tracking, 3D Gaussian training \& segmentation and finally, 3D style transfer. In the first step, we utilize off-the-shelf image segmentation and tracking models to obtain consistent 2D masks across the complete scene to generate temporally coherent mask identifiers. Then, these masks are used as supervision for segmenting the 3D Gaussians into distinct objects while jointly optimizing their geometry and color, allowing for accurate object selection. Finally, we use a nearest-neighbor feature matching loss to finetune the selected Gaussians by aligning their spherical harmonic coefficients with the provided style image to achieve consistency and visual appeal. This method provides accurate and focused stylization, resulting in more customized and high-quality outcomes.

Our method produces visually pleasing results across diverse scenes and datasets, highlighting its versatility with different artistic styles.

\section{Related Work}
Style transfer on 3D scenes has attracted attention in recent years, with notable contributions emerging for both NeRF and 3DGS representations. Recently, several techniques have also been introduced for object editing within the 3DGS framework. Here, we review these prior efforts to provide context for our proposed approach. 

\subsection{Style Transfer}
Image style transfer, a long-standing challenge in computer vision, involves optimizing a content image and a reference style image to create a new image. The resulting image maintains the content of the original while adopting the style of the reference. Early works \cite{gatys2015neuralalgorithmartisticstyle, neuralstyletransfer} utilize VGG-Net \cite{simonyan2015deepconvolutionalnetworkslargescale} to extract multi-level features and iteratively optimize a Gram matrix loss and a content loss to render a new image. Since then, numerous works have been proposed to improve various aspects of style transfer, including faster stylization \cite{huang2017arbitrarystyletransferrealtime, johnson2016perceptuallossesrealtimestyle, 9711417} and improving semantic consistency and texture preservation \cite{gu2018arbitrarystyletransferdeep, kolkin2019styletransferrelaxedoptimal, liao2017visualattributetransferdeep}. Several approaches have been proposed to constrain the style transfer to specific objects in the image. \cite{castillo2017sonzornslemmatargeted} simultaneously segments and stylizes individual objects in an image and uses a Markov random field for anti-aliasing near the object boundaries for seamless blending. \cite{kurzman2019classbasedstylingrealtimelocalized} performs localized style transfer in real-time by generating object masks using a fast semantic segmentation deep neural network. \cite{article} utilizes Segment Anything (SAM) \cite{kirillov2023segment} to generate object masks and contain the style transfer to specific objects and uses a novel loss function to ensure smooth boundaries between the stylized and non-stylized parts of the image.

\subsection{Style Transfer in Radiance Fields}
Several works have been introduced to extend stylization to 3D scenes. ARF \cite{zhang2022arf} presents a method for transferring artistic styles to 3D scenes represented by neural radiance fields. Their approach effectively incorporates artistic features into rendered images while maintaining multi-view consistency by using a nearest neighbor feature matching (NNFM) loss for 3D stylization instead of the Gram matrix loss. However, ARF applies style changes to the entire scene, modifying all objects within the view, which might not be desirable in applications where style transfer is intended for specific objects. Building upon ARF, several approaches were introduced to localize style transfer to user-specified objects. ARF-Plus \cite{li2023arfpluscontrollingperceptualfactors} introduces perceptual controls for scale, spatial area, and perceived depth and uses semantic segmentation masks to ensure that stylization is applied only to selected areas. S2RF \cite{lahiri2023s2rf} leverages Plenoxel's \cite{yu_and_fridovichkeil2021plenoxels} grid based representation for the 3D scene and a masked NNFM (mNNFM) loss to constrain the style transfer only to desired areas. CoARF \cite{coarf} extends S2RF by adding a semantic component to style transfer using LSeg \cite{lseg} features in addition to VGG features. However, these methods suffer from slow rendering speeds due to their reliance on ray marching.  In the 3DGS paradigm, StyleGaussian \cite{liu2024stylegaussian} and GSS \cite{gsinstyle} introduce approaches for scene stylization. Both methods seek to generate novel views of a scene using unseen style images at test time, after being trained on a large dataset of style images. Similar to ARF \cite{zhang2022arf}, these approaches are limited to stylizing the entire scene.

\subsection{Object Editing in 3D Gaussian Splatting}
A common approach for localized editing in Gaussian splatting involves appending additional features to each Gaussian to encode semantic information. These features are optimized by rendering feature maps similar to RGB rasterization and using 2D feature maps or segmentation masks from foundation models as guidance. GaussianEditor \cite{GaussianEditor} adds a scalar feature to each Gaussian to identify whether the Gaussian is in the editing region of interest (ROI). It then uses 2D grounded segmentation masks to optimize this feature. Gaussian Grouping \cite{ye2023gaussian} appends a feature vector to each Gaussian and uses masks from SAM \cite{kirillov2023segment} and DEVA \cite{cheng2023tracking} as guidance. Similarly, Feature 3DGS \cite{feature3dgs} distills LSeg \cite{lseg} and SAM features into each Gaussian for promptable editing. All these approaches demonstrate text-guided editing as a downstream task. However, textual descriptions in style transfer can be ambiguous, making it difficult to accurately specify particular colors, styles, or textures within 3D scenes. Several diffusion-based methods have also been proposed. Instruct-GS2GS \cite{igs2gs} utilizes a 2D diffusion model to modify the appearance of 3D objects with text instructions but it fails to constrain the changes to the specified object. TIP-Editor \cite{tipeditor} personalizes a diffusion model using LoRA and accepts both an image and a text prompt with a 3D bounding box for local editing. Fine-tuning typically requires 5 to 25 minutes with this approach. By contrast, our method is lightweight and achieves results in less than a minute.

\subsection{Concurrent Work}
Several interesting works have emerged that focus on localized image-conditioned editing of 3D Gaussians. One such work is StylizedGS \cite{stylizedgs}, which emphasizes scene stylization while allowing for spatial control through the use of 2D masks, allowing different styles for different regions. However, since the approach uses only 2D masks, it faces significant limitations. Since each Gaussian has a volume, alpha blending may inadvertently include neighboring Gaussians in the computation graph during rasterization, causing unintended style transfer to other parts of the scene. This results in the inability to precisely stylize a single object while leaving the rest of the scene unchanged. Another notable work is ICE-G \cite{iceg}. Unlike our method, ICE-G copies the style image to the ROI for a single 2D view. It then employs SAM and DINO \cite{dino} to propagate the style to multiple views, effectively updating the data set with the desired style. Fine-tuning for style transfer is subsequently performed on this updated dataset. 

\section{Method} 
\begin{figure}[!htb]
    \centerline{\includegraphics[width=1\columnwidth]{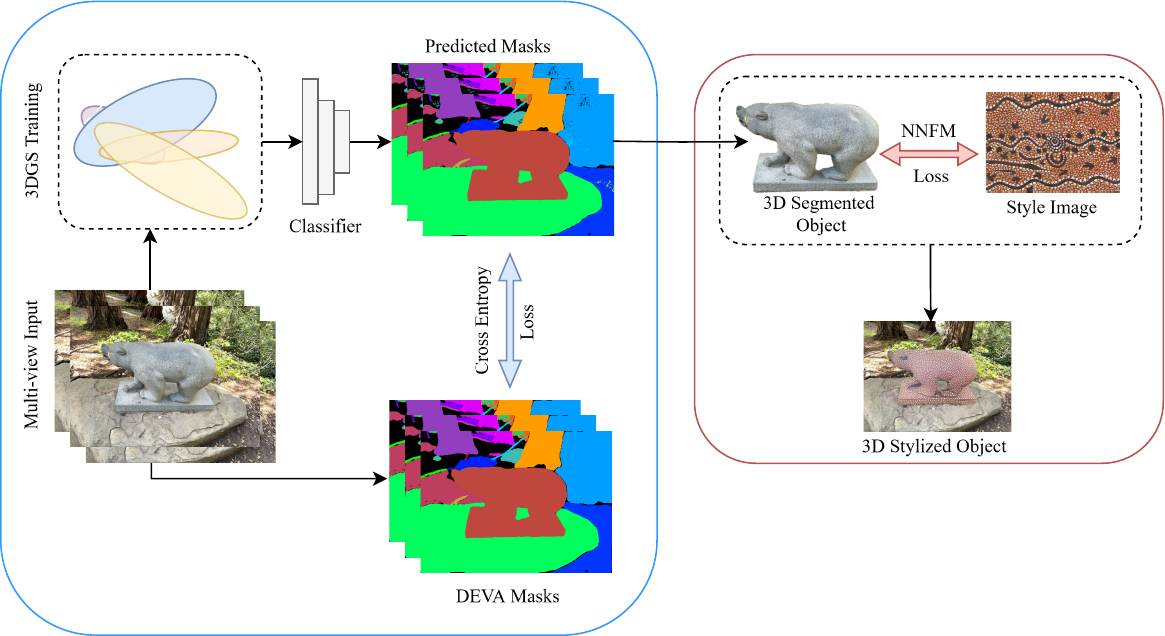}}
    \caption{Our approach for StyleSplat. We first use an off-the-shelf segmentation and tracking model \cite{cheng2023tracking} to generate view-consistent 2D object masks. Then, we use the multi-view images to learn the geometry and color of 3D Gaussians while simultaneously learning a per Gaussian feature vector. These feature vectors are decoded into object labels using a linear classifier to collect the Gaussians corresponding to the user-specified objects. The SH coefficients of these selected Gaussians are finetuned to align with the style image using NNFM loss.}
    \label{fig:method}
\end{figure}

Given a reference style image, a set of posed images, and objects specified by the user, we aim to achieve fast novel view synthesis such that the objects corresponding to the user input are stylized according to the reference style image. Our approach involves three steps: 2D mask generation \& object tracking, 3D Gaussian training \& segmentation, and 3D style transfer. \cref{fig:method} provides a brief overview of our method.

\subsection{Preliminary: 3D Gaussian Splatting}

3D Gaussians \cite{Kerbl20233DGS} is an explicit 3D scene representation. Each 3D Gaussian is characterized by a covariance matrix $\Sigma$ and a center point $\mu$ , which is referred to as the mean value of the Gaussian as:
\begin{equation}
    G(\mathbf{x}) = e^{-\frac{1}{2} \textit{$\mathbf{x}$}^{T} \Sigma^{-1} \textit{$\mathbf{x}$} }
\end{equation}

For differentiable optimization, the covariance matrix $\Sigma$ can be decomposed into a scaling matrix S and a rotation matrix:    
\begin{equation}
    \Sigma = \textbf{$RSS^{T}R^{T}$}
\end{equation}
% R:
% Σ = RSST RT
% . (2)
When rendering novel views, differential splatting is employed for the 3D Gaussians within the camera planes. As introduced by \cite{surfacesplat}, using a viewing transform matrix W and the Jacobian matrix J of the affine approximation of the projective transformation, the covariance matrix $ \Sigma' $ in camera coordinates can be computed as:

\begin{equation}
    \Sigma {}' = JW \Sigma W^{T}J^{T}
\end{equation}

In summary, each 3D Gaussian is characterized by the following attributes: position $X \in R^3$, color defined by spherical harmonic (SH) coefficients $C \in R^k$
(where k represents number of SH coefficents), opacity $\alpha \in R$, rotation factor $r \in R^4$, and scaling factor $s \in R^3$. Specifically, for each pixel, the color and opacity of all the Gaussians are computed using the Gaussian’s representation Eq. 1. The blending of N ordered points that overlap the pixel is given by the formula:

\begin{equation}
    C = \Sigma c_{i} \alpha_{i} \prod^{i-1}_{j=1} (1-\alpha_{i})
\end{equation}

Here, $c_{i} , \alpha_i$ represents the density and color of this point
computed by a 3D Gaussian G with covariance $\Sigma$ multiplied by an optimizable per-point opacity and SH color coefficients.

\subsection{2D Mask Generation \& Object Tracking} \label{2D_mask}
Before segmenting the 3D scene, we need to acquire accurate 2D segmentation masks on the entire sequence. These masks need to be temporally coherent to ensure that the different class indices correspond to the same object across frames. We treat the captured images as a video sequence and make use of a DEcoupled Video segmentation Approach (DEVA) \cite{cheng2023tracking} with a strong zero-shot segmentation model (SAM) \cite{kirillov2023segment} to get temporally coherent masks.

\subsection{3D Gaussian Training \& Segmentation} \label{3dgs_seg}
We follow the approach of Gaussian Grouping \cite{ye2023gaussian} to jointly train and segment 3D Gaussians. More specifically, each 3D Gaussian is given a learnable compact feature vector of length 16. These encodings are optimized similar to the spherical harmonic coefficients in the 3DGS pipeline. For a given view, the feature vector for a single pixel is evaluated as follows:
\[
E_{id} = \sum_{i \in N} e_i \alpha_i' \prod_{j=i}^{i-1}(1-\alpha_j') 
\]
where $e_i'$ is the feature vector for the $i$th Gaussian, and $\alpha_i'$ is the influence of the $i$th Gaussian on the current pixel, evaluated similar to \cite{Yifan_2019}. The rendered $E_{id}$ for each pixel is then passed through a classifier to provide a class label. A cross-entropy loss is used between the predicted class labels vs the class labels obtained in the first stage. Additionally, a spatial consistency loss is added which ensures that the feature vectors for the top-k nearest 3D Gaussians are similar. Once this stage is completed, all Gaussians in the scene corresponding to the same object have similar feature vectors.

\subsection{3D Style Transfer} \label{3d_style}
Once we have obtained a scene representation with segmented 3D Gaussians, we use the learned feature vectors to perform style transfer on user-specified objects. To get the mask IDs of specific objects in the scene, the user can specify a bounding box around the object or utilize Grounding DINO \cite{liu2023grounding} to extract the IDs using a text prompt. We select the Gaussians corresponding to the objects of interest by passing the feature vectors through the trained classifier and filtering out Gaussians with activations less than a threshold. We also perform statistical outlier removal, eliminating Gaussians whose positions deviate significantly from their neighbors compared to the average for the scene. We then freeze all the properties of the selected Gaussians and enable gradients only for their SH coefficients. For each training view, we apply the nearest neighbor feature matching (NNFM) loss between the VGG features of the rendered image and the reference style image. The NNFM loss minimizes the cosine distance between the VGG feature of each pixel in the render with its nearest neighbor in the style image and is given by:
\[
L_{NNFM} = \frac{1}{N}\sum_{i} \min_{j} \left( F_r(i) \cdot F_s(j) \right)
\]
where $F_r$ are the VGG features for the render, $F_s$ are the features for the style image, and $N$ is the number of pixels in the rendered image. Since only the SH coefficients of the user-specified object are trainable, the style transfer is contained to a single object.

\section{Results}

\begin{figure}[!htb]
    \centering
    \begin{subfigure}[b]{0.3\linewidth}
        \centering
        \includegraphics[width=\linewidth]{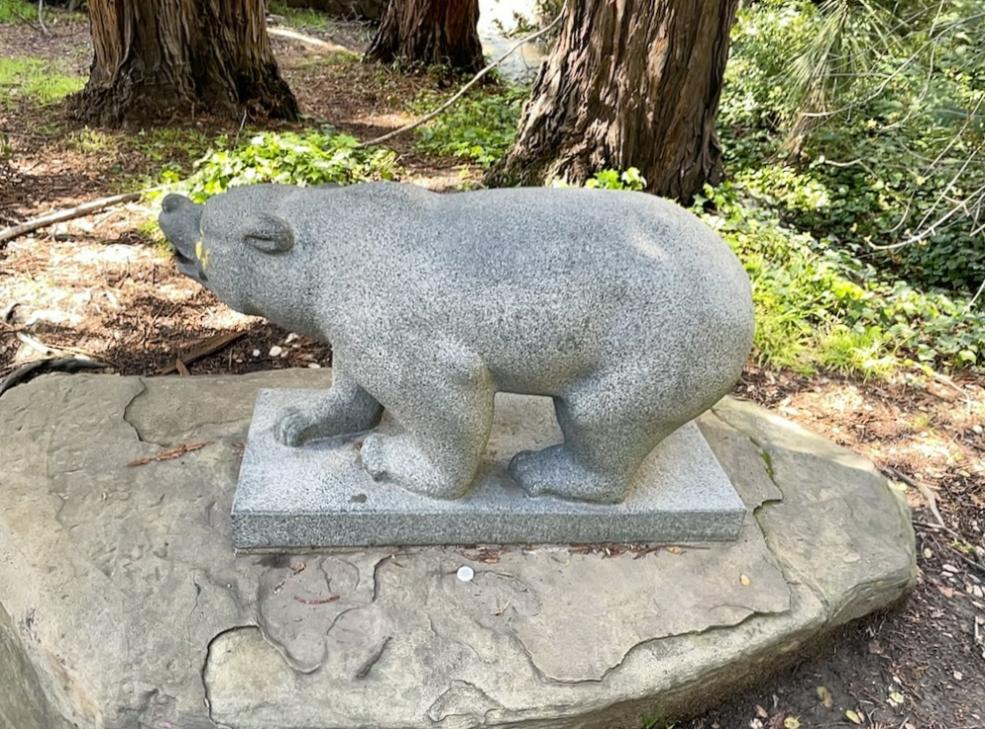}
    \end{subfigure}
    \begin{subfigure}[b]{0.3\linewidth}
        \centering
        \includegraphics[width=\linewidth]{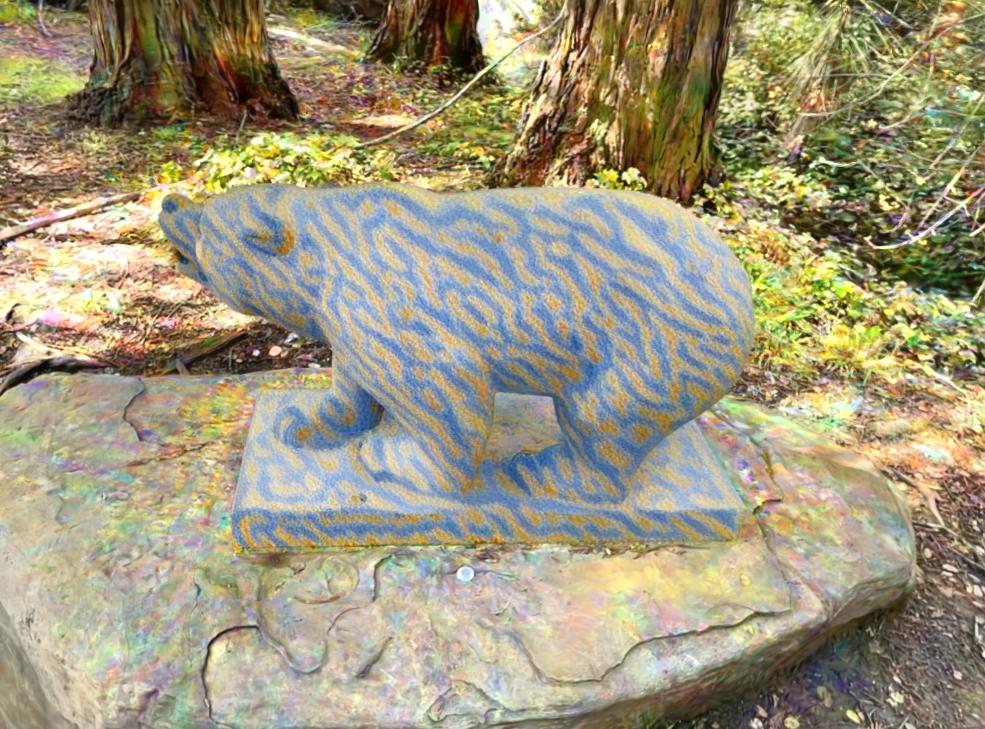}
    \end{subfigure}
    \begin{subfigure}[b]{0.3\linewidth}
        \centering
        \includegraphics[width=\linewidth]{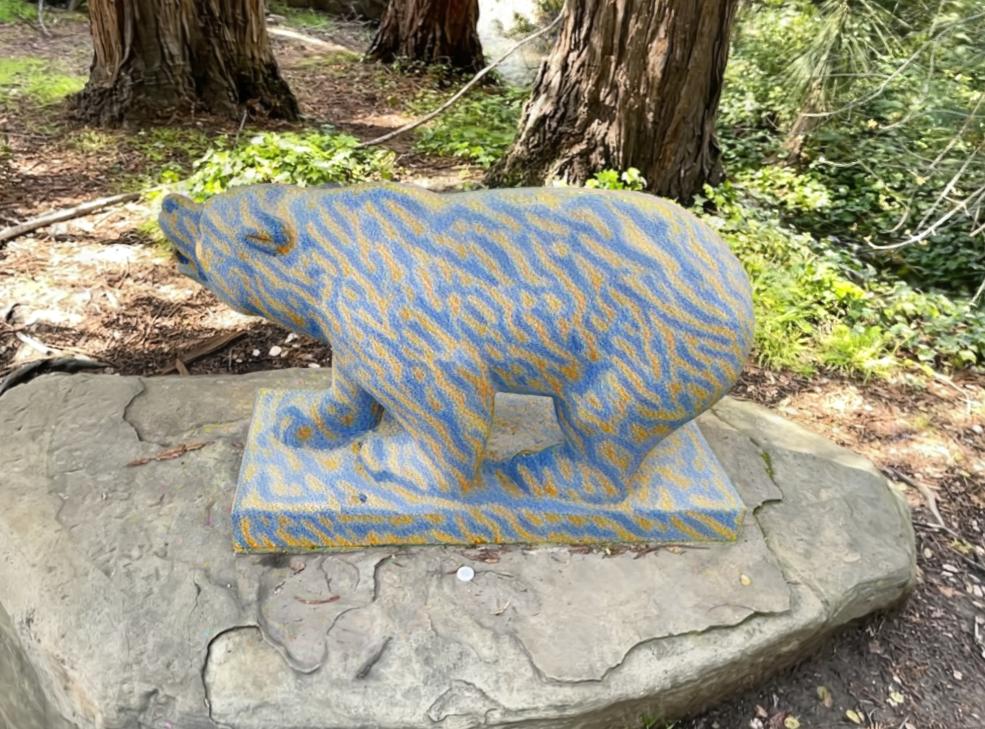}
    \end{subfigure}
    \begin{subfigure}[b]{0.3\linewidth}
        \centering
        \includegraphics[width=\linewidth]{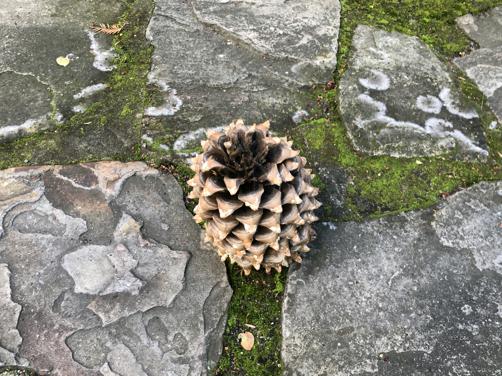}
        \caption{Ground Truth}
    \end{subfigure}
    \begin{subfigure}[b]{0.3\linewidth}
        \centering
        \includegraphics[width=\linewidth]{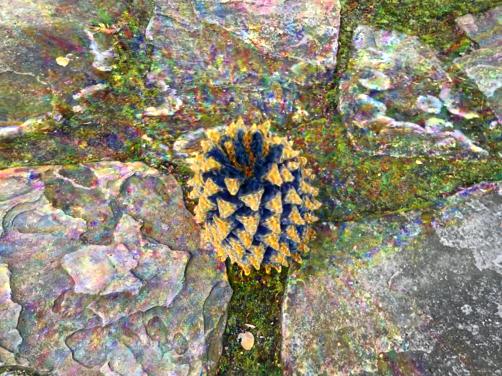}
        \caption{2D Segmentation}
    \end{subfigure}
    \begin{subfigure}[b]{0.3\linewidth}
        \centering
        \includegraphics[width=\linewidth]{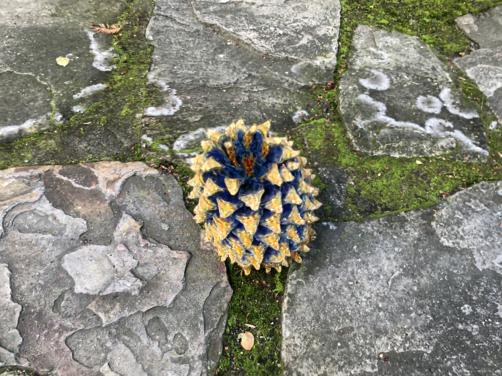}
        \caption{3D Segmentation}
    \end{subfigure}
    \caption{Effect of 3D segmentation on localized style transfer. The first column shows the initial 3D object. The second column demonstrates the limitations of using a masked loss similar to previous radiance field-based approaches \cite{lahiri2023s2rf, coarf, li2023arfpluscontrollingperceptualfactors}. 2D masks can be inconsistent across views and introduce errors, leading to artifacts in different parts of the scene due to incorrect Gaussians being modified. The third column illustrates the benefits of training with a collection of noisy masks to learn a view-consistent feature vector per Gaussian, effectively correcting these errors and avoiding leakage.}
    \label{fig:leakage}
\end{figure}

To assess our method, we perform qualitative evaluations on multiple real-world scenes. We visually demonstrate the effectiveness of our method by successfully applying different styles to various objects in a diverse selection of scenes. This section highlights how our 3D segmentation approach for localized stylization prevents leakages (\cref{objselect}), its performance in both single-object (\cref{singleobj}) and multi-object settings (\cref{multiobj}), and scale control (\cref{subsec:control}). Finally, we qualitatively compare our approach with S2RF (\cref{subsec:s2rf}).

\subsection{Implementation Details} 

We evaluate our approach on a range of real-world scenes from diverse datasets, including NeRF \cite{mildenhall2020nerfrepresentingscenesneural}, MipNerf360 \cite{mipnerf}, LERF \cite{lerflanguageembeddedradiance}, and InstantNGP \cite{mueller2022instant}. These datasets provide various challenging environments to test the effectiveness of our method. Additionally, we use style images from the WikiArt dataset \cite{wikiartWikiArtorgVisual}, which offers a wide variety of artistic styles, demonstrating the versatility of our style transfer technique.

For all scenes, we begin by running the 3D Gaussian and segmentation pipeline for an initial 30,000 iterations. Following this, we freeze the parameters of all the Gaussians, restricting further optimization to only the spherical harmonic (SH) coefficients of the Gaussians that correspond to the selected object. The style transfer optimization is then performed using 25\% of the training images, running for 500 to 1,000 iterations depending on the complexity of the scene. This targeted optimization process is highly efficient, taking less than a minute to complete on a single NVIDIA A100 GPU.

\subsection{Object selection} \label{objselect} Previous radiance field based approaches \cite{lahiri2023s2rf, coarf, li2023arfpluscontrollingperceptualfactors} use a 2D masked loss in the image space to localize the style transfer. However, 2D masks can be inconsistent across views and contain errors, leading to incorrect Gaussians being stylized (as demonstrated in \cref{fig:leakage}). Although this is not a problem in neural scene representations, this manifests as artifacts in the final stylized scene for 3DGS. Using a 3D segmentation approach leads to robustness against 2D mask errors.

\begin{figure}[!t]
    \centering
    \begin{subfigure}[b]{0.18\linewidth}
        \centering
        \includegraphics[width=\linewidth]{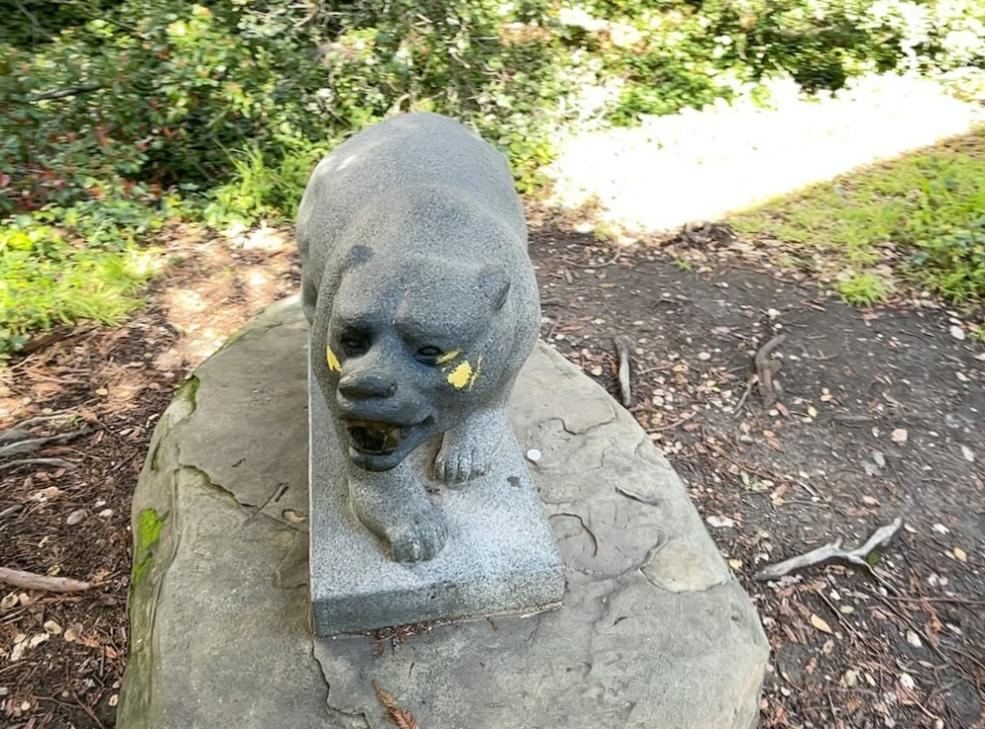}
        \caption*{}
    \end{subfigure}
    \begin{subfigure}[b]{0.18\linewidth}
        \centering
        \includegraphics[width=\linewidth]{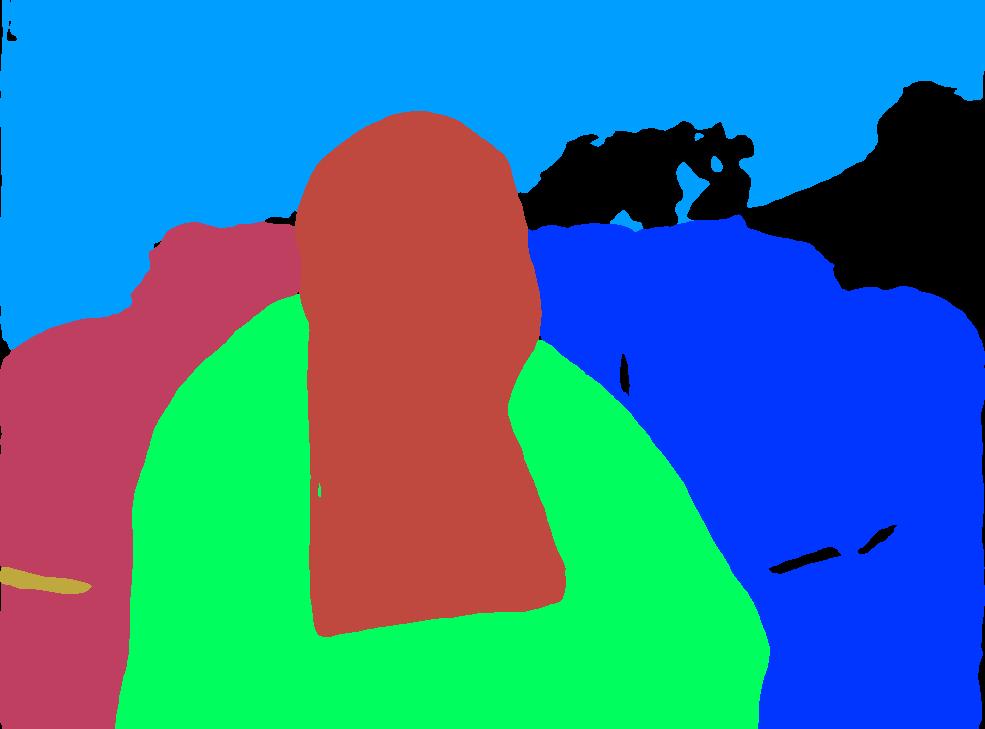}
        \caption*{}
    \end{subfigure}
    \begin{subfigure}[b]{0.18\linewidth}
        \centering
        \includegraphics[width=\linewidth]{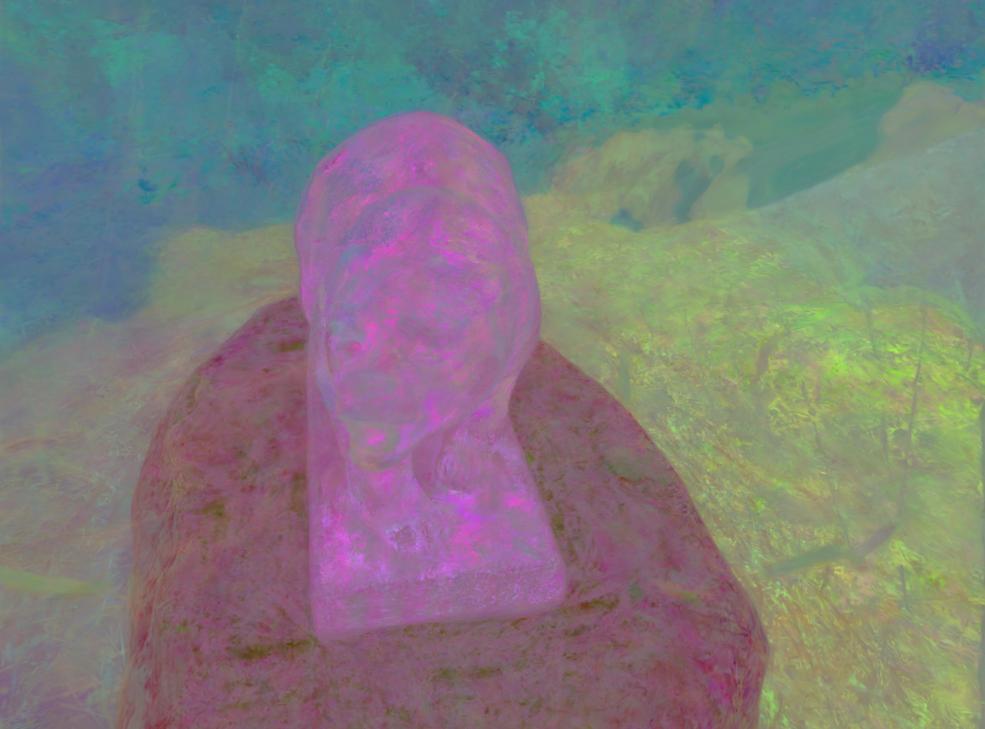}
        \caption*{}
    \end{subfigure}
    \begin{subfigure}[b]{0.18\linewidth}
        \centering
        \includegraphics[width=\linewidth]{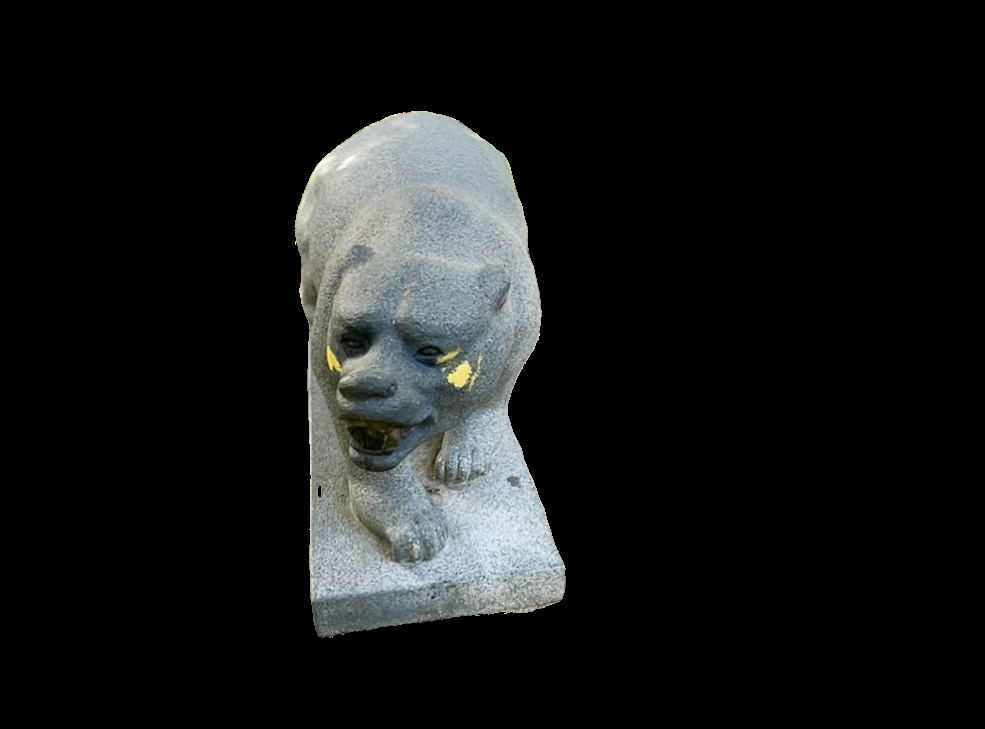}
        \caption*{}
    \end{subfigure}
    \begin{subfigure}[b]{0.18\linewidth}
        \centering
        \includegraphics[width=\linewidth]{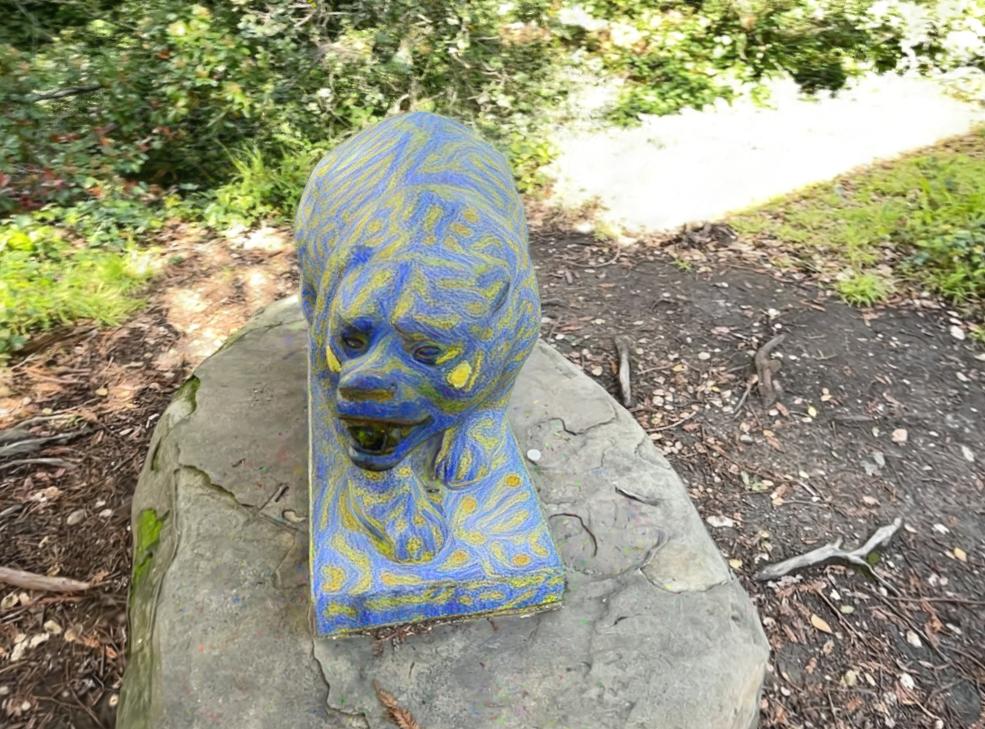}
        \caption*{}
    \end{subfigure}
    \begin{subfigure}[b]{0.18\linewidth}
        \centering
        \includegraphics[width=\linewidth]{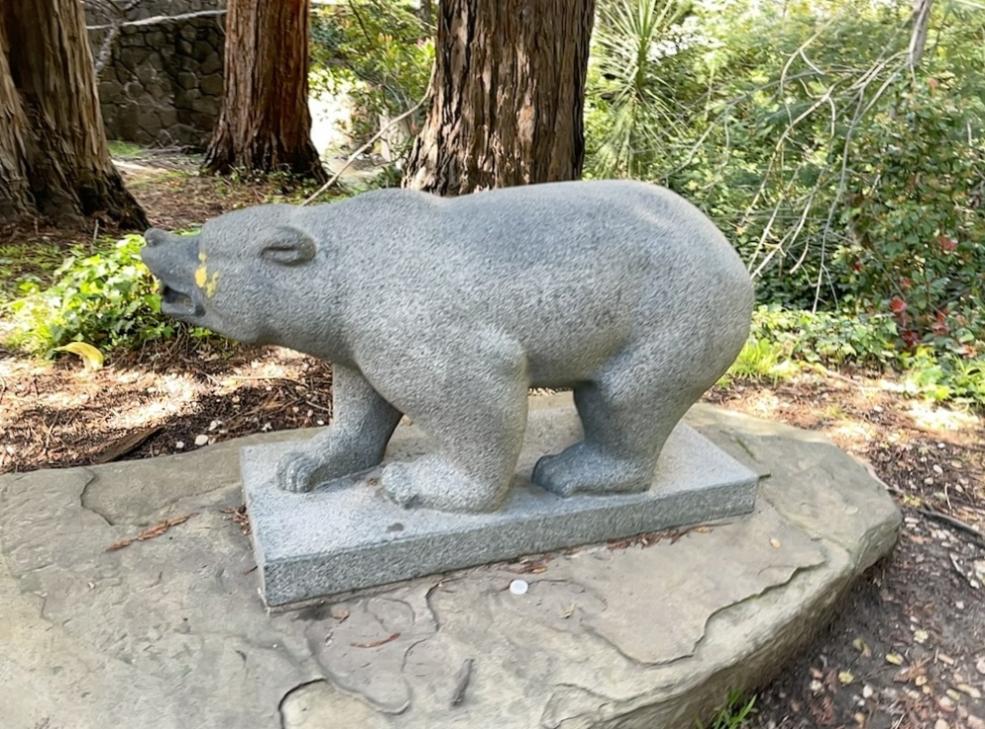}
        \caption*{}
    \end{subfigure}
    \begin{subfigure}[b]{0.18\linewidth}
        \centering
        \includegraphics[width=\linewidth]{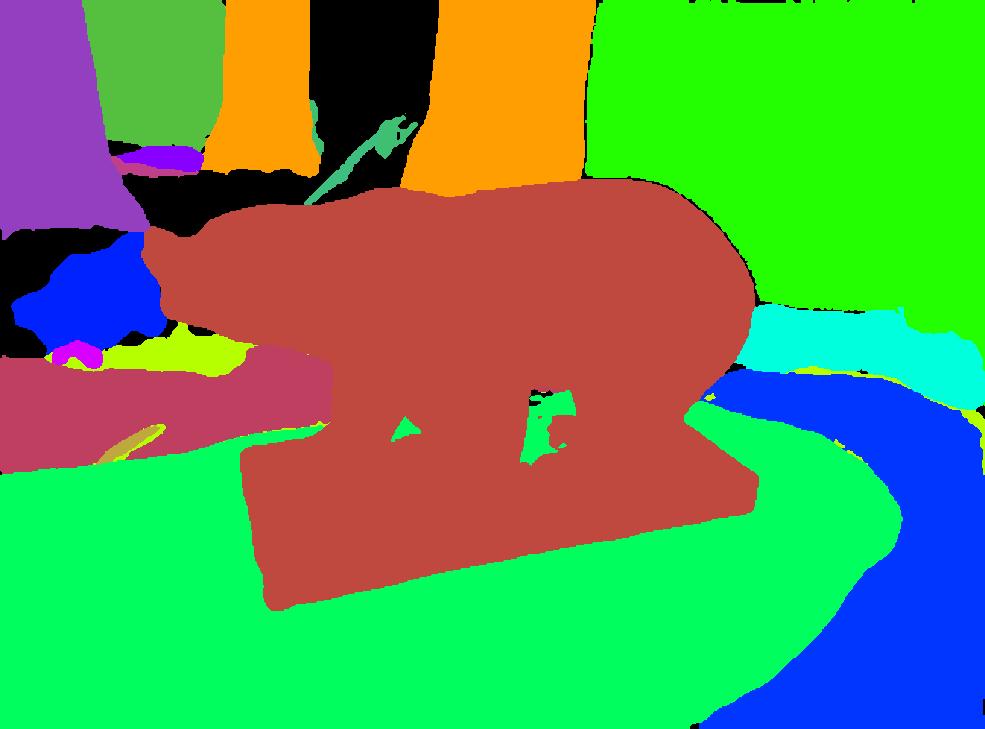}
        \caption*{}
    \end{subfigure}
    \begin{subfigure}[b]{0.18\linewidth}
        \centering
        \includegraphics[width=\linewidth]{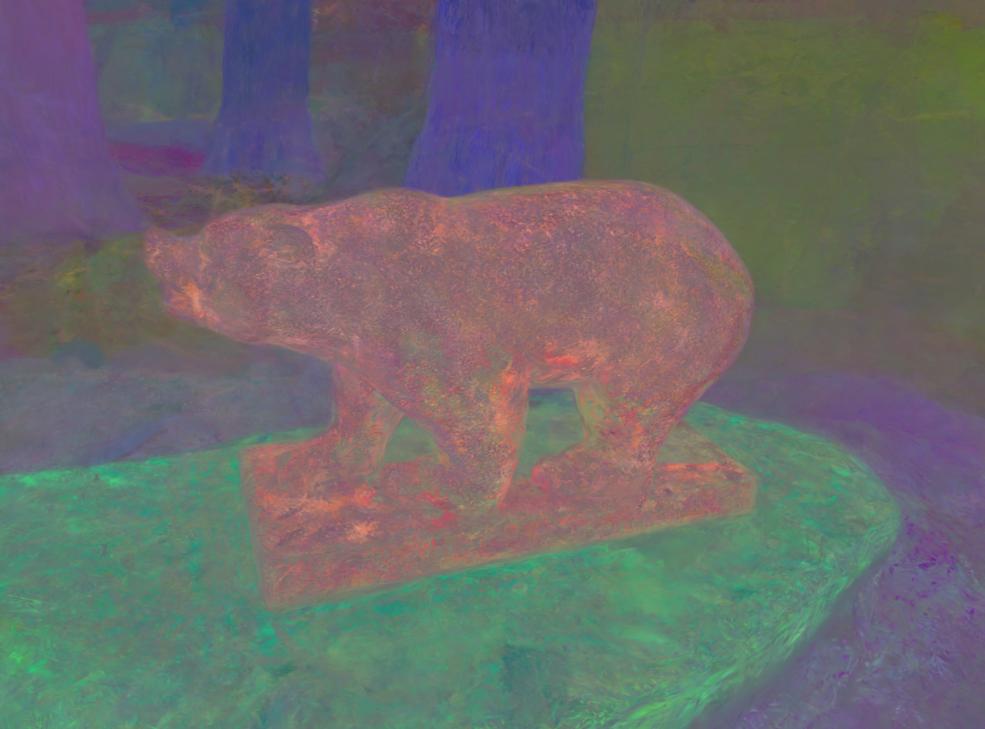}
        \caption*{}
    \end{subfigure}
    \begin{subfigure}[b]{0.18\linewidth}
        \centering
        \includegraphics[width=\linewidth]{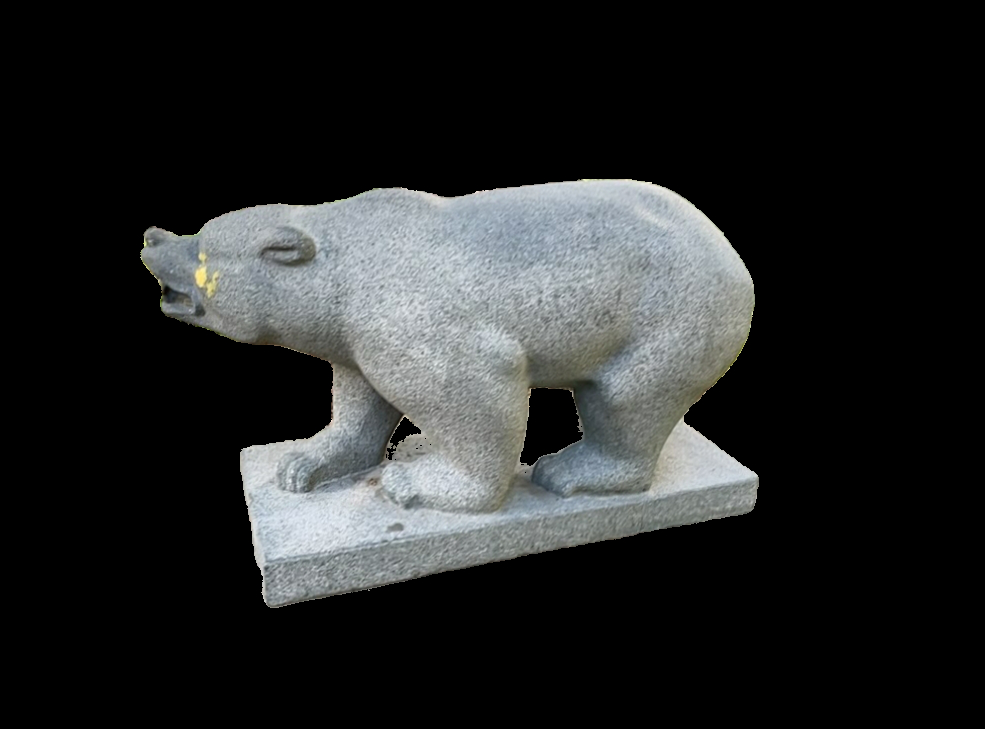}
        \caption*{}
    \end{subfigure}
    \begin{subfigure}[b]{0.18\linewidth}
        \centering
        \includegraphics[width=\linewidth]{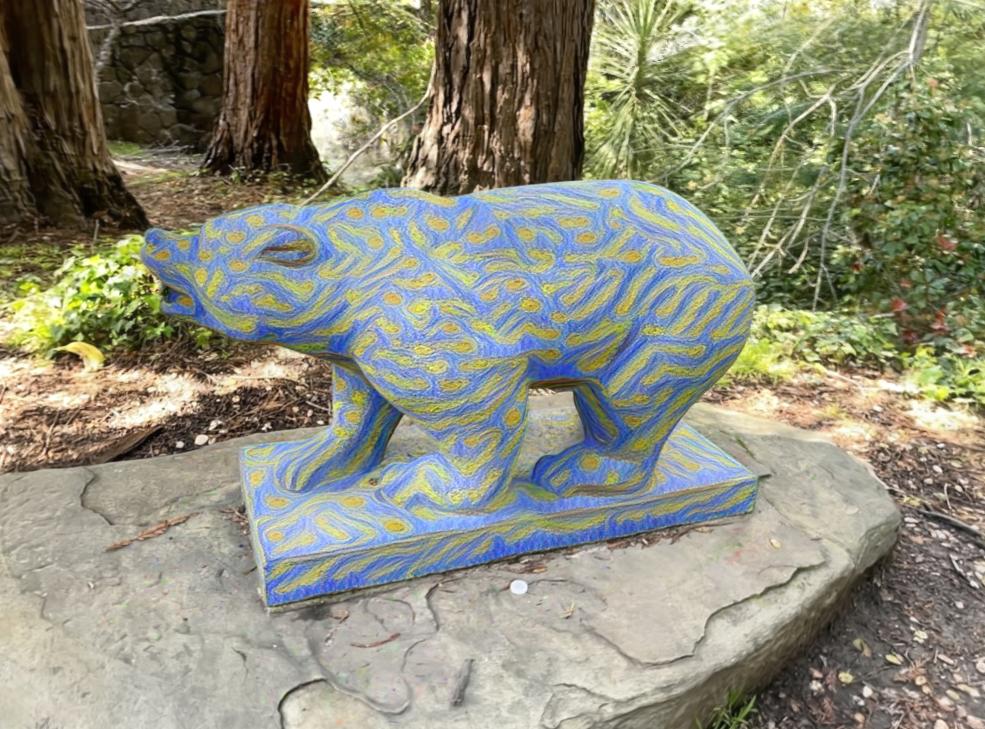}
        \caption*{}
    \end{subfigure}
    \begin{subfigure}[b]{0.18\linewidth}
        \centering
        \includegraphics[width=\linewidth]{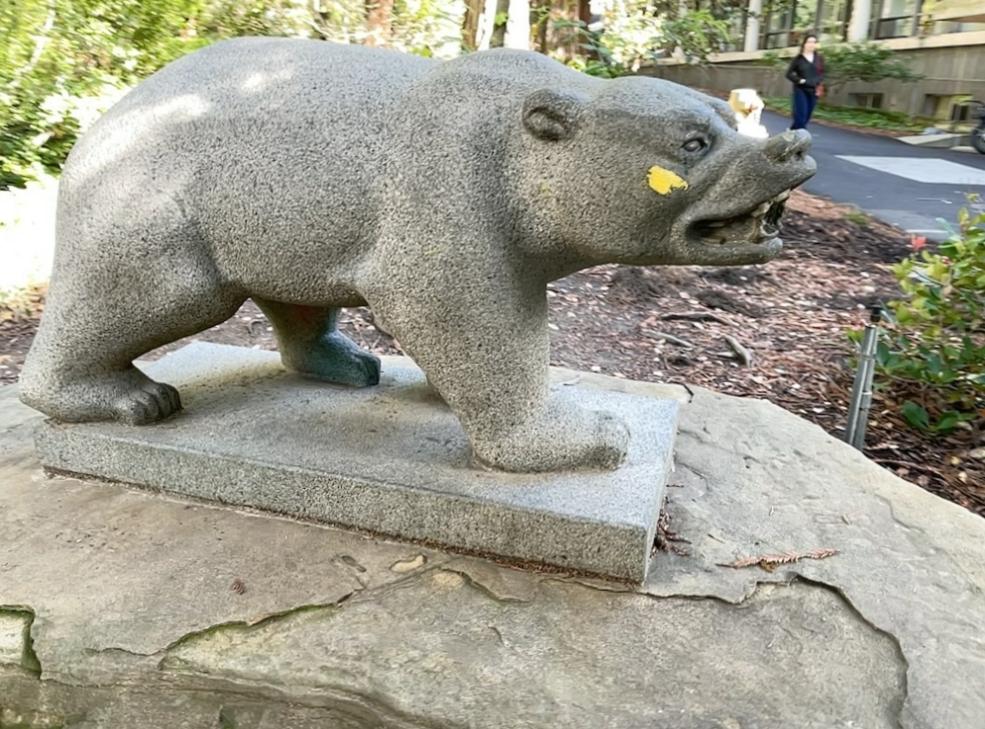}
        \caption*{}
    \end{subfigure}
    \begin{subfigure}[b]{0.18\linewidth}
        \centering
        \includegraphics[width=\linewidth]{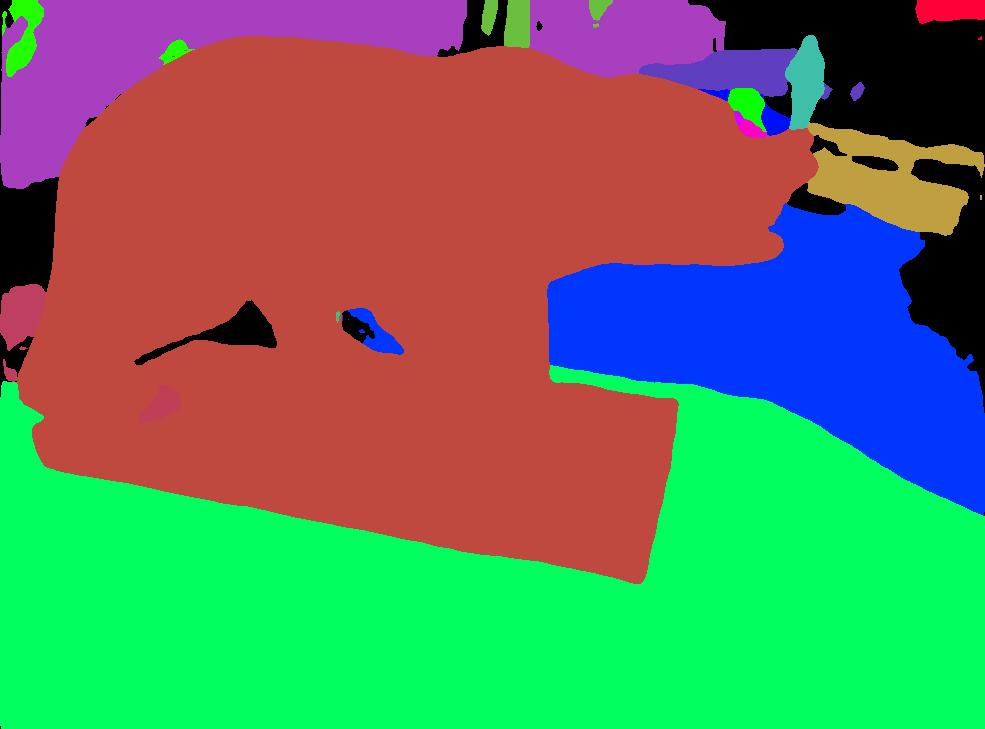}
        \caption*{}
    \end{subfigure}
    \begin{subfigure}[b]{0.18\linewidth}
        \centering
        \includegraphics[width=\linewidth]{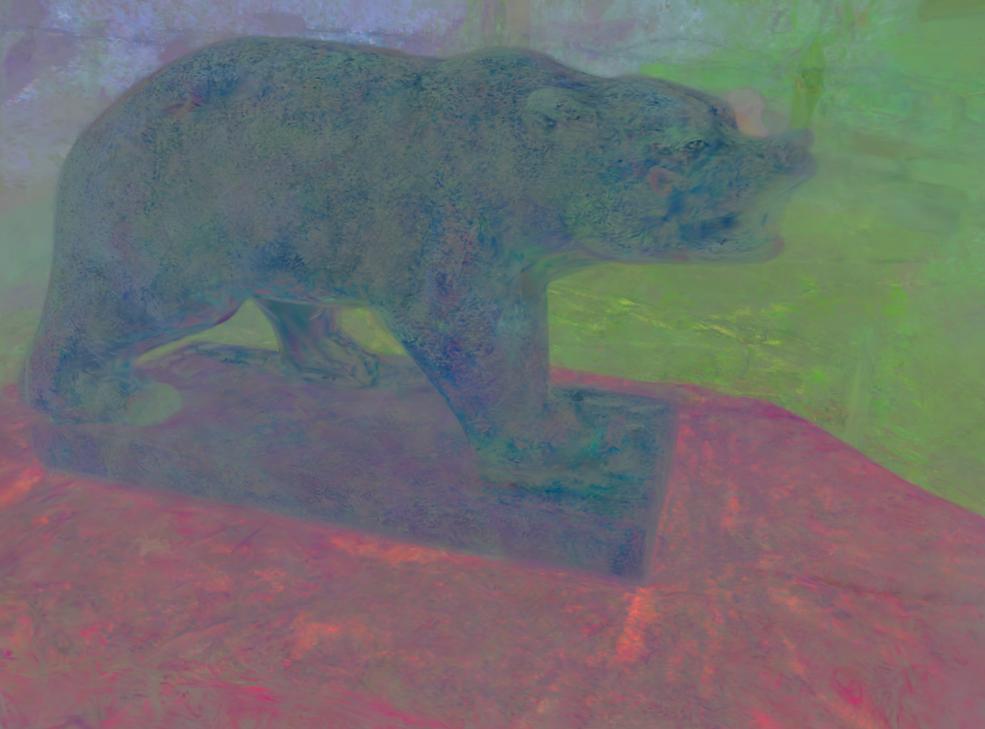}
        \caption*{}
    \end{subfigure}
    \begin{subfigure}[b]{0.18\linewidth}
        \centering
        \includegraphics[width=\linewidth]{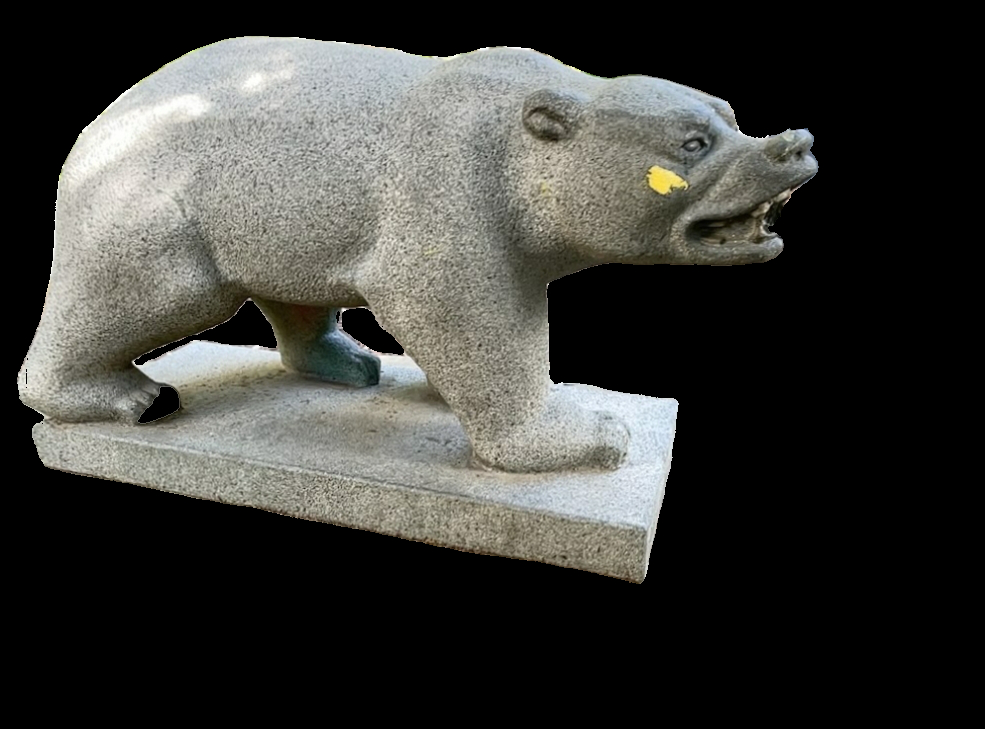}
        \caption*{}
    \end{subfigure}
    \begin{subfigure}[b]{0.18\linewidth}
        \centering
        \includegraphics[width=\linewidth]{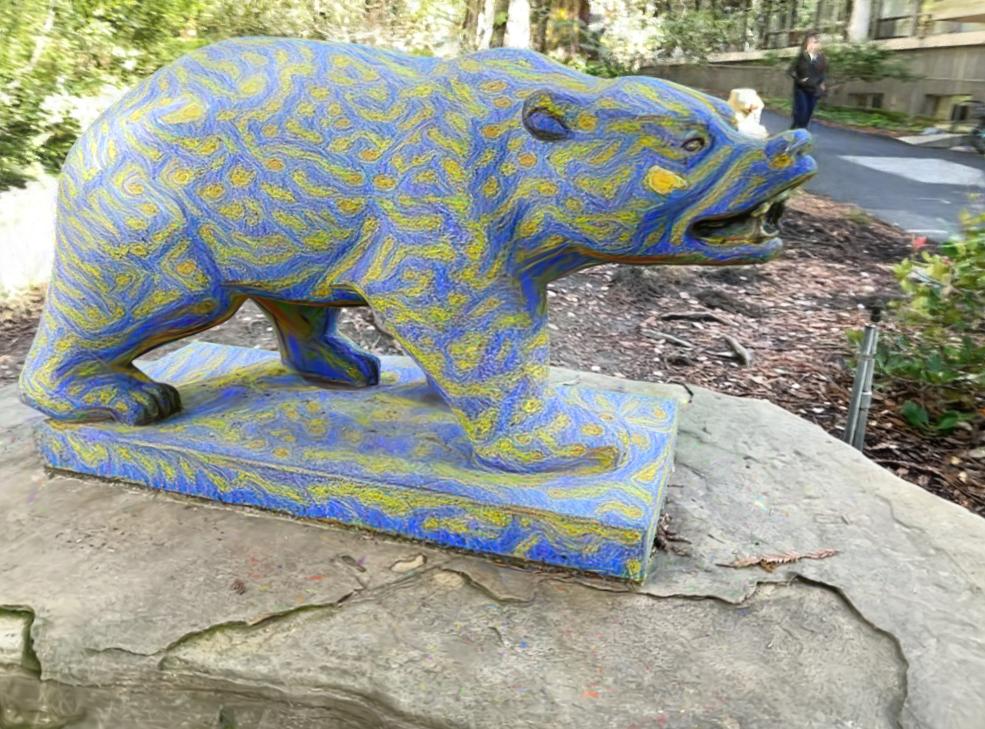}
        \caption*{}
    \end{subfigure}
    \begin{subfigure}[b]{0.18\linewidth}
        \centering
        \includegraphics[width=\linewidth]{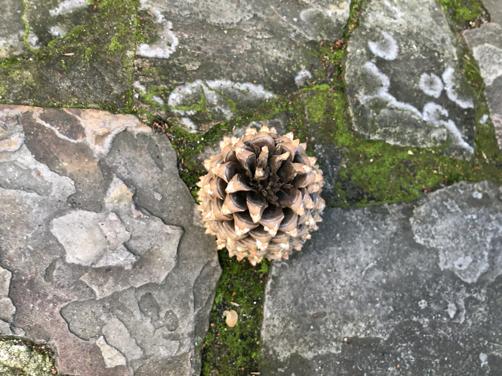}
        \caption*{}
    \end{subfigure}
    \begin{subfigure}[b]{0.18\linewidth}
        \centering
        \includegraphics[width=\linewidth]{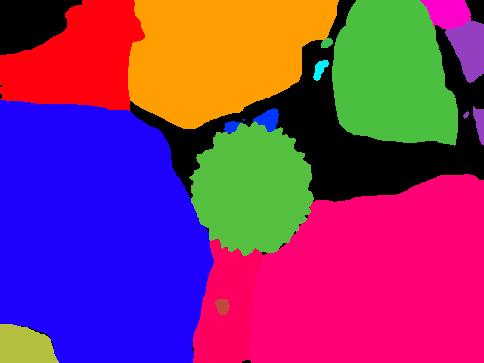}
        \caption*{}
    \end{subfigure}
    \begin{subfigure}[b]{0.18\linewidth}
        \centering
        \includegraphics[width=\linewidth]{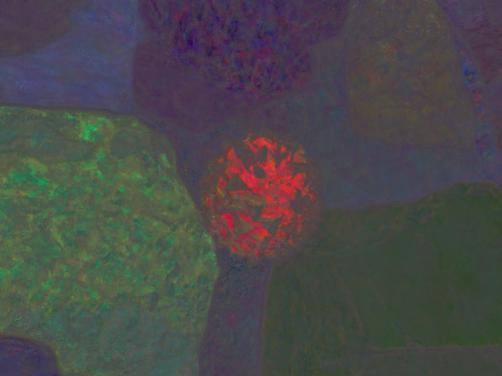}
        \caption*{}
    \end{subfigure}
    \begin{subfigure}[b]{0.18\linewidth}
        \centering
        \includegraphics[width=\linewidth]{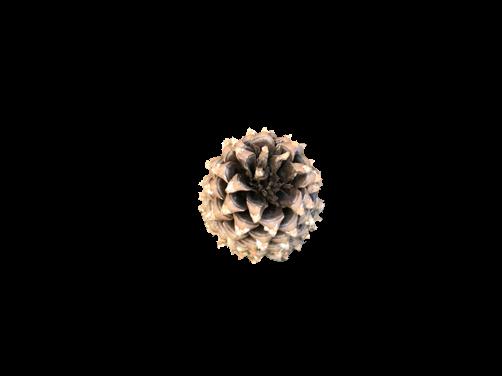}
        \caption*{}
    \end{subfigure}
    \begin{subfigure}[b]{0.18\linewidth}
        \centering
        \includegraphics[width=\linewidth]{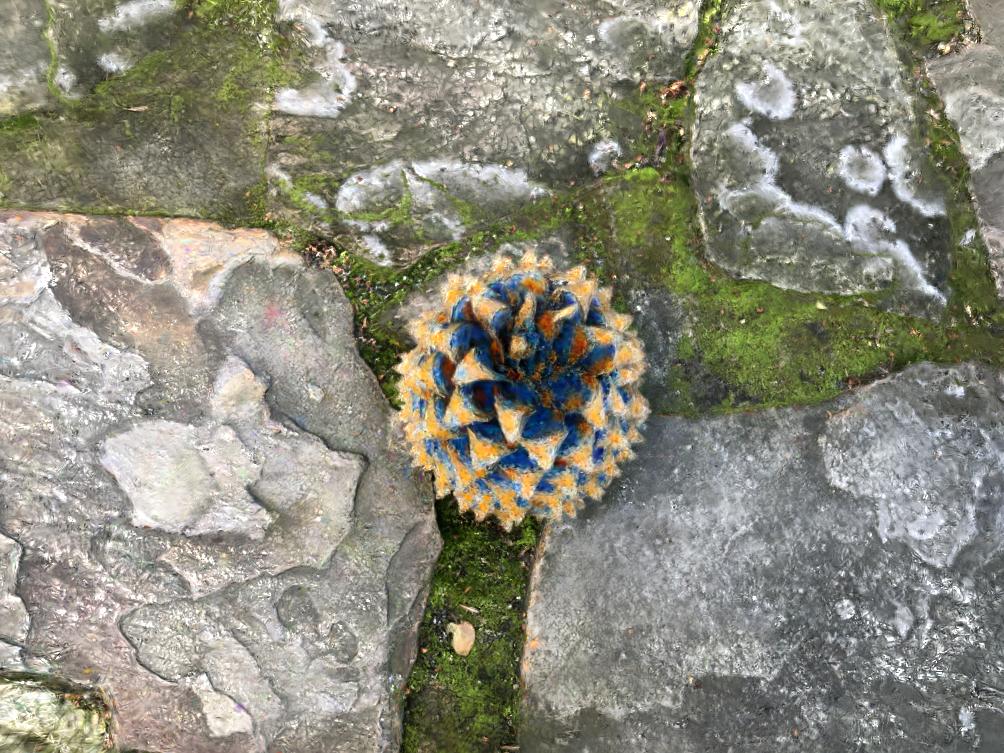}
        \caption*{}
    \end{subfigure}
    \begin{subfigure}[b]{0.18\linewidth}
        \centering
        \includegraphics[width=\linewidth]{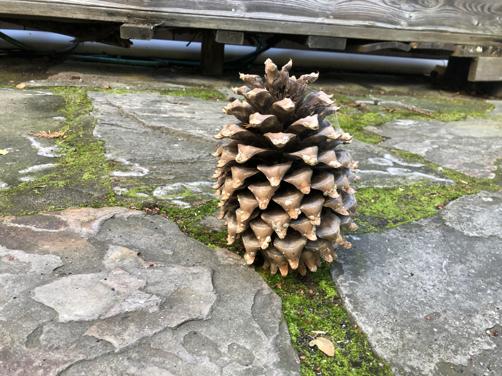}
        \caption*{}
    \end{subfigure}
    \begin{subfigure}[b]{0.18\linewidth}
        \centering
        \includegraphics[width=\linewidth]{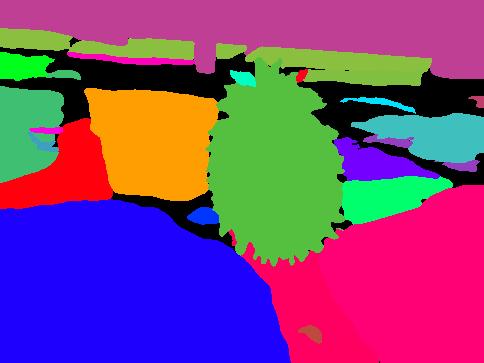}
        \caption*{}
    \end{subfigure}
    \begin{subfigure}[b]{0.18\linewidth}
        \centering
        \includegraphics[width=\linewidth]{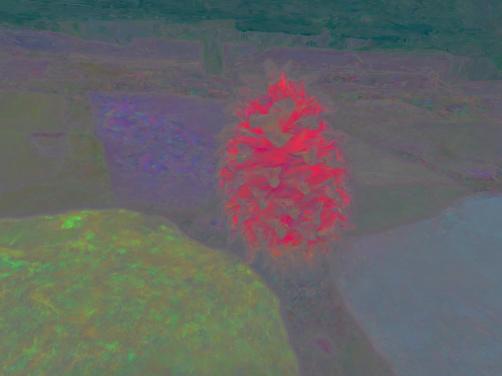}
        \caption*{}
    \end{subfigure}
    \begin{subfigure}[b]{0.18\linewidth}
        \centering
        \includegraphics[width=\linewidth]{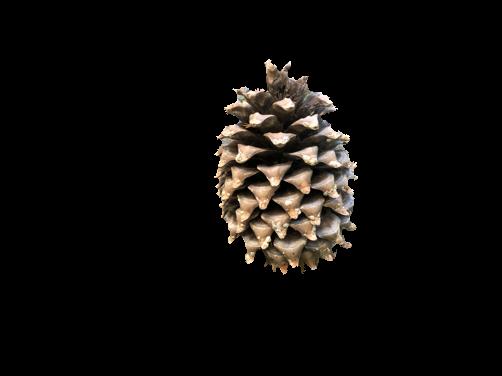}
        \caption*{}
    \end{subfigure}
    \begin{subfigure}[b]{0.18\linewidth}
        \centering
        \includegraphics[width=\linewidth]{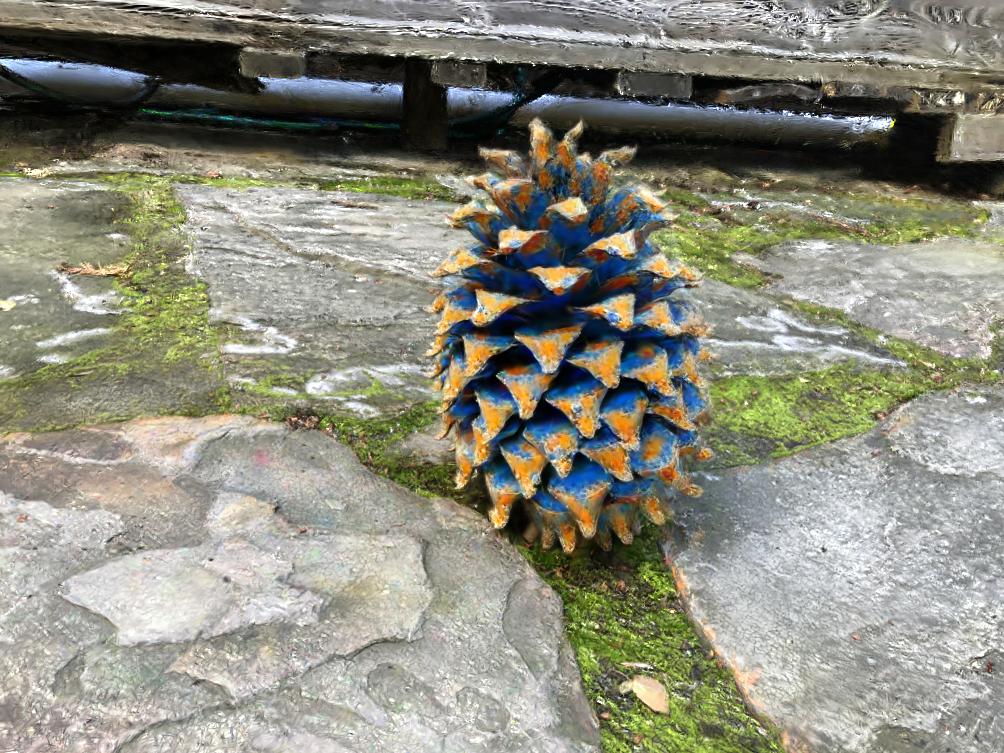}
        \caption*{}
    \end{subfigure}
    \begin{subfigure}[b]{0.18\linewidth}
        \centering
        \includegraphics[width=\linewidth]{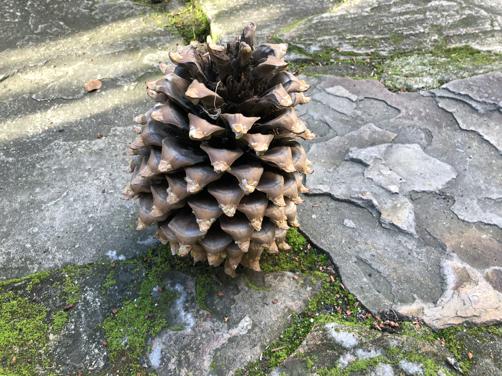}
        \caption{Ground Truth}
    \end{subfigure}
    \begin{subfigure}[b]{0.18\linewidth}
        \centering
        \includegraphics[width=\linewidth]{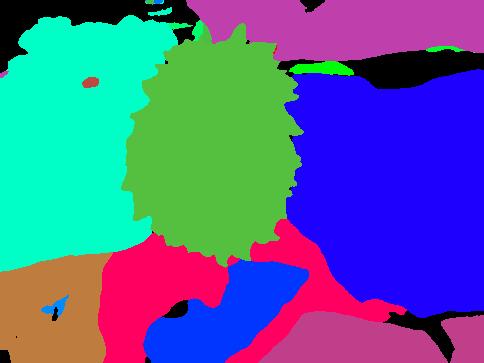}
        \caption{DEVA Masks}
        \label{fig:bear_gt_mask}
    \end{subfigure}
    \begin{subfigure}[b]{0.18\linewidth}
        \centering
        \includegraphics[width=\linewidth]{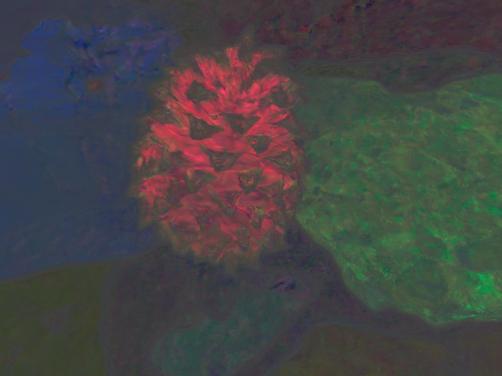}
        \caption{Features}
        \label{fig:bear_pred_feats}
    \end{subfigure}
    \begin{subfigure}[b]{0.18\linewidth}
        \centering
        \includegraphics[width=\linewidth]{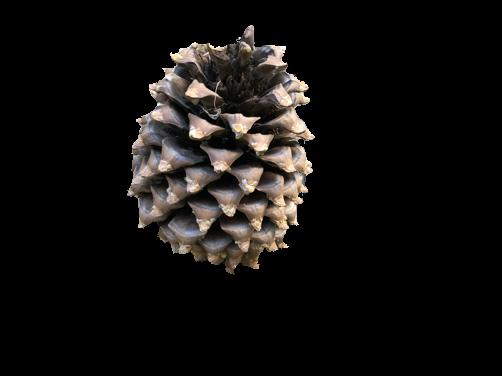}
        \caption{Object}
        \label{fig:bear_pred_mask}
    \end{subfigure}
    \begin{subfigure}[b]{0.18\linewidth}
        \centering
        \includegraphics[width=\linewidth]{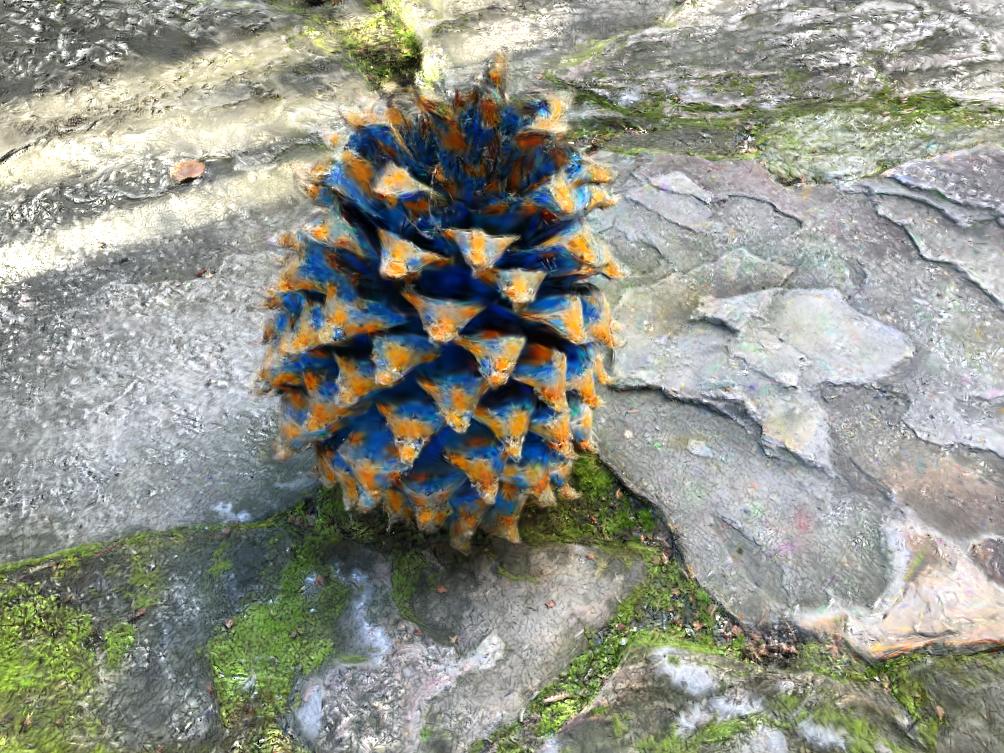}
        \caption{Stylized}
    \end{subfigure}

    \caption{3D segmentation results. 
    Figure \textbf{(a)} shows the ground truth image, \textbf{(b)} displays the masks extracted using SAM and DEVA, \textbf{(c)} visualizes the learned feature vectors of all objects in the scene, \textbf{(d)} presents the extracted object, and \textbf{(e)} illustrates the final stylized result.}

    \label{fig:bear}
\end{figure}

The results of our object selection approach are illustrated in \cref{fig:bear}. The masks provided by DEVA are shown in \cref{fig:bear_gt_mask} and \cref{fig:bear_pred_feats} visualizes the learnt per-Gaussian feature vectors for the \textit{bear} and \textit{pinecone} scenes from the InstantNGP and NeRF datasets respectively. The feature vectors are visualized as the first three principal component analysis (PCA) components of the original 16-dimensional vectors. We can observe that the approach provides an effective way to select 3D objects in the scene, confining the style transfer to the selected object. 

\subsection{Single object style transfer} \label{singleobj} 
\begin{figure}[!tb]
    \centering
    \begin{tabular}{c c c c}
        \multicolumn{4}{c}{} \\

        & \includegraphics[width=2cm, height=1.6cm]{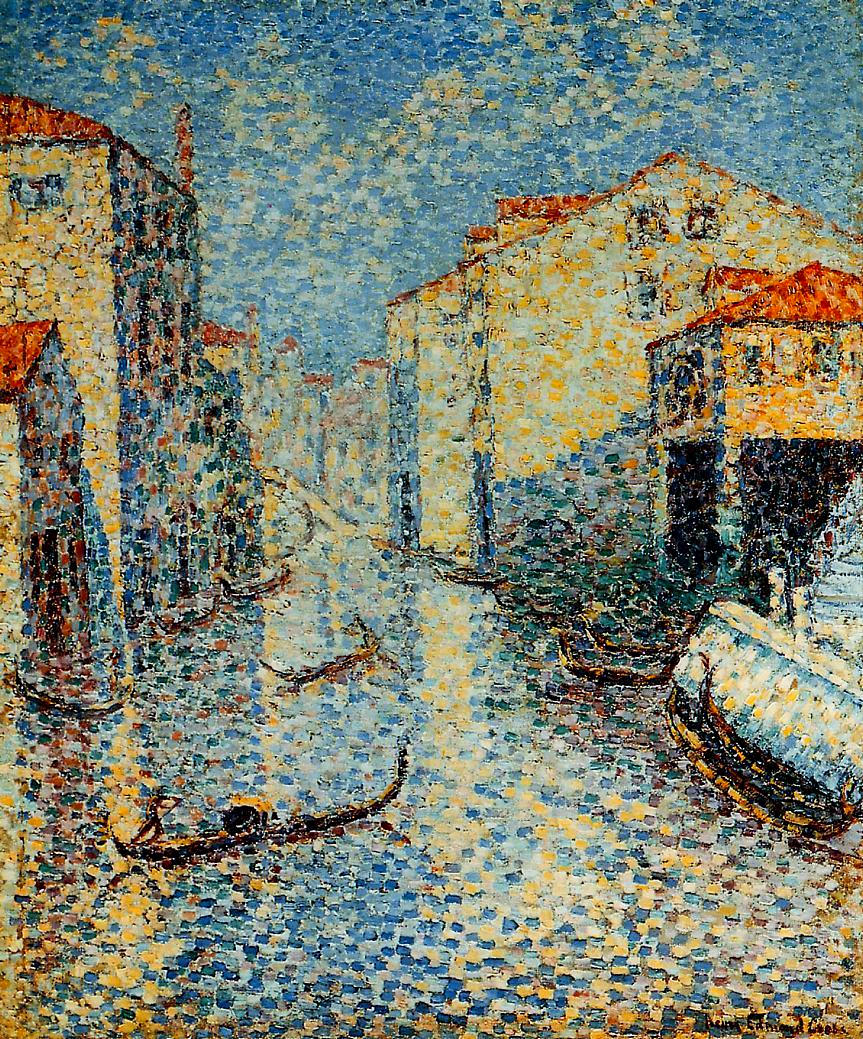} & \includegraphics[width=2cm, height=1.6cm]{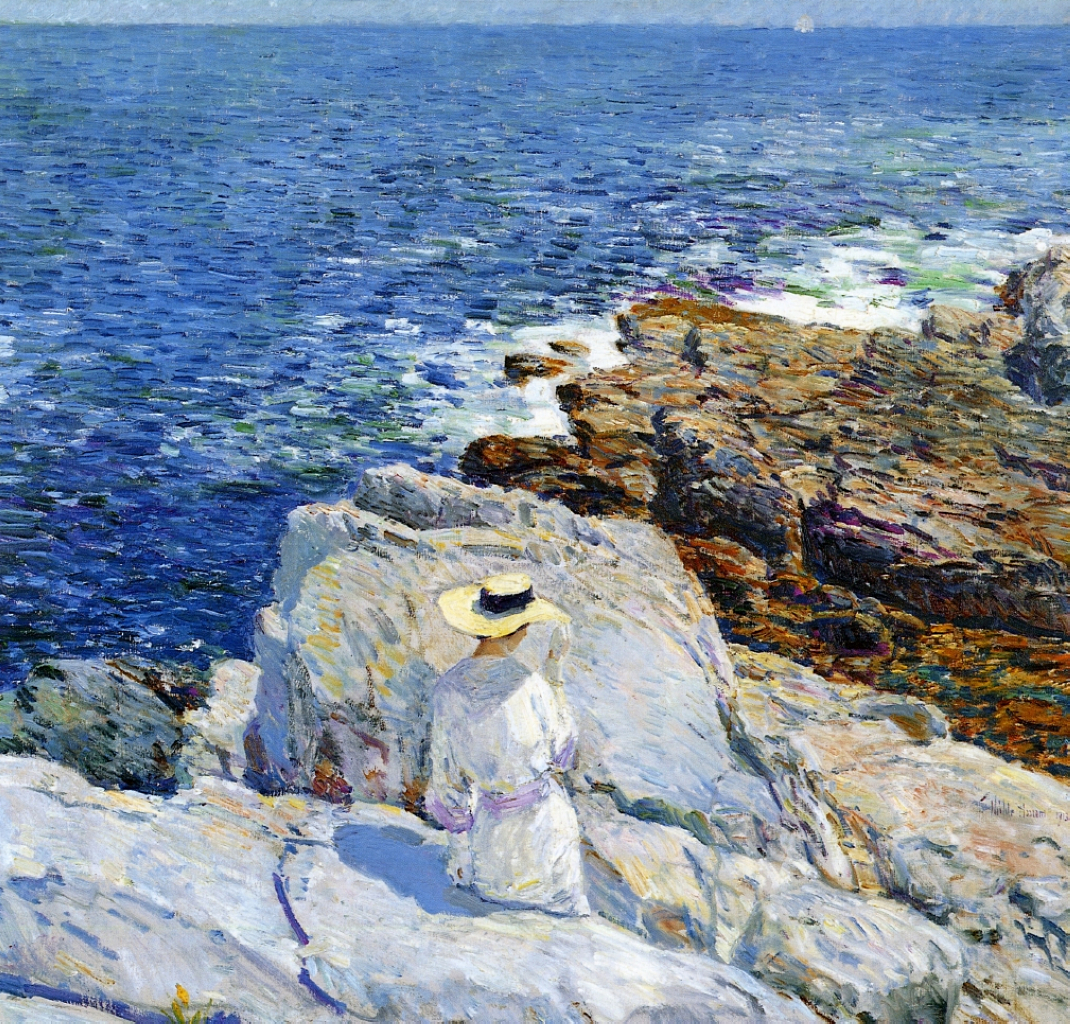} & \includegraphics[width=2cm, height=1.6cm]{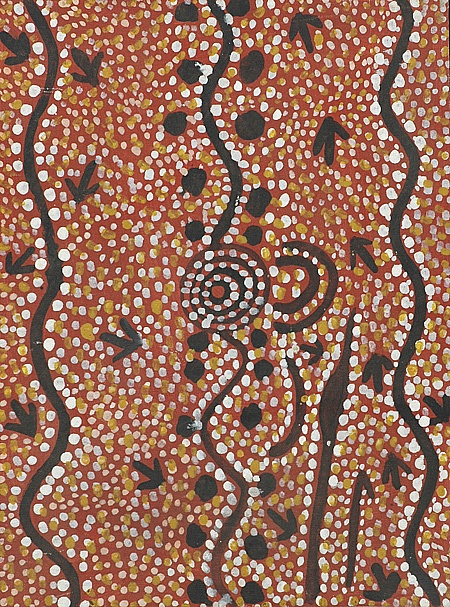} \\
        % &  \\
        
        \includegraphics[width=2.75cm]{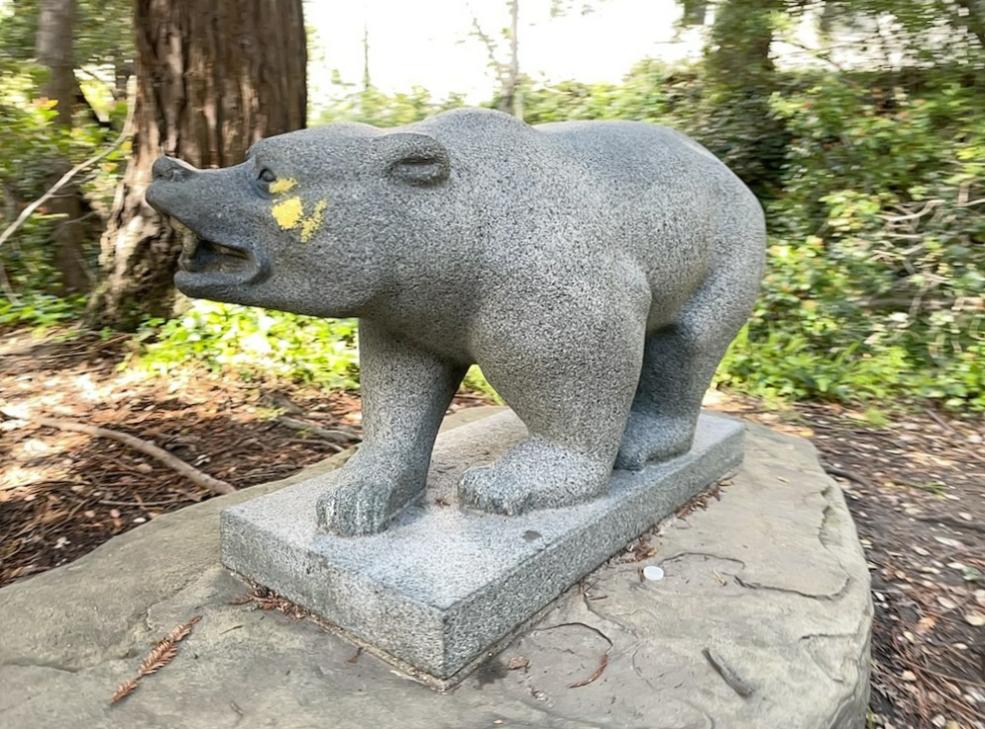} & \includegraphics[width=2.75cm]{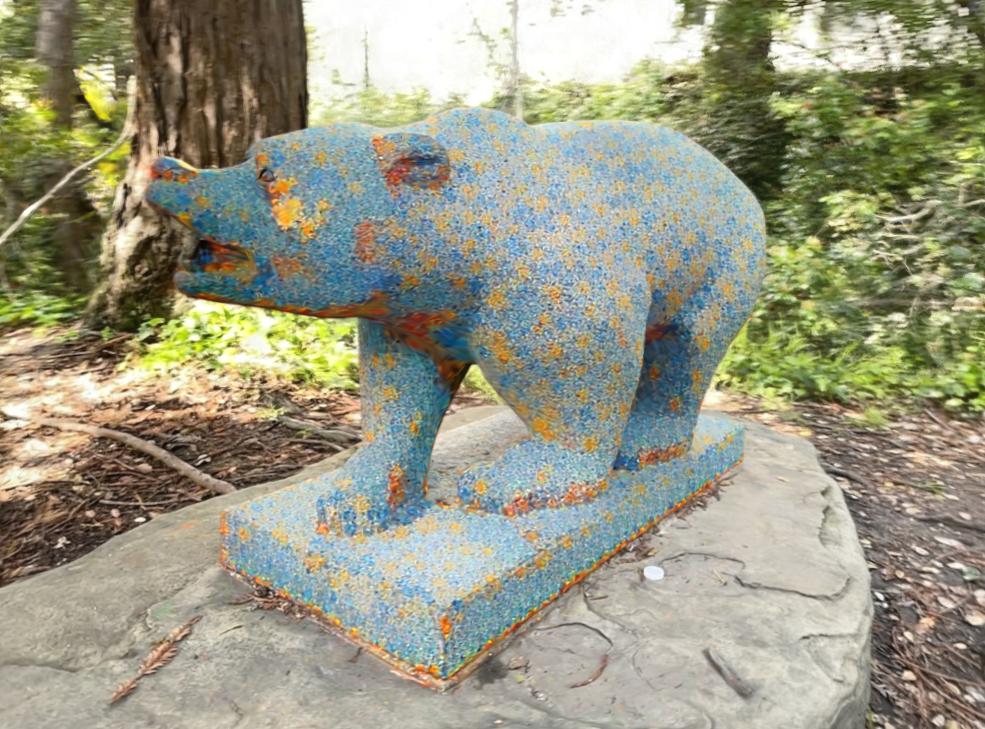} & \includegraphics[width=2.75cm]{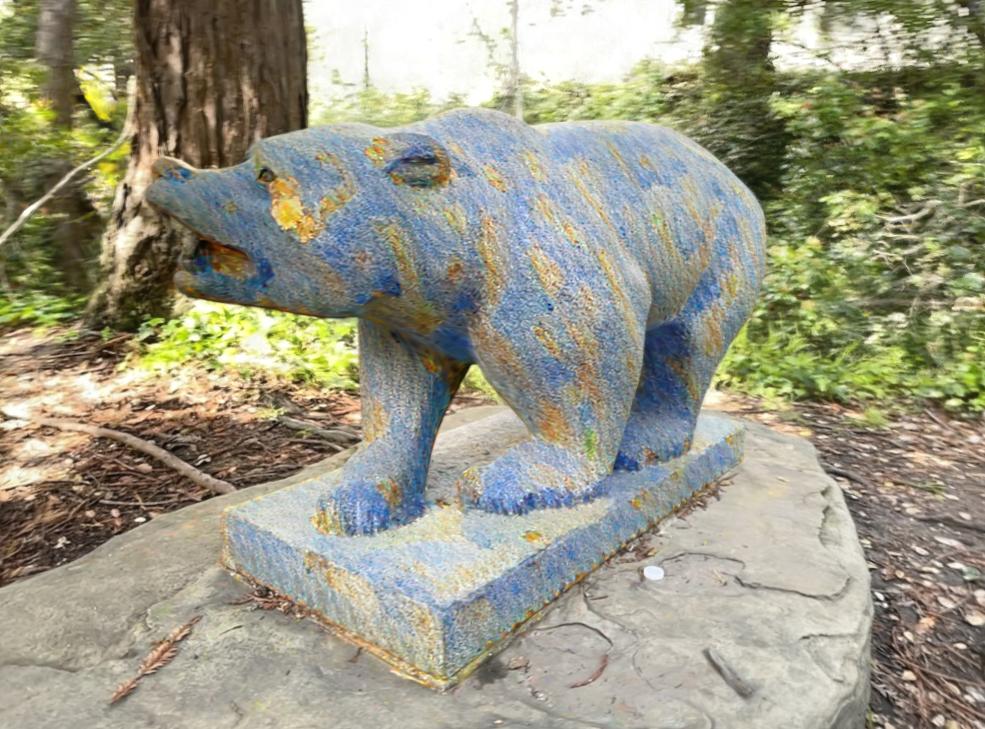} & \includegraphics[width=2.75cm]{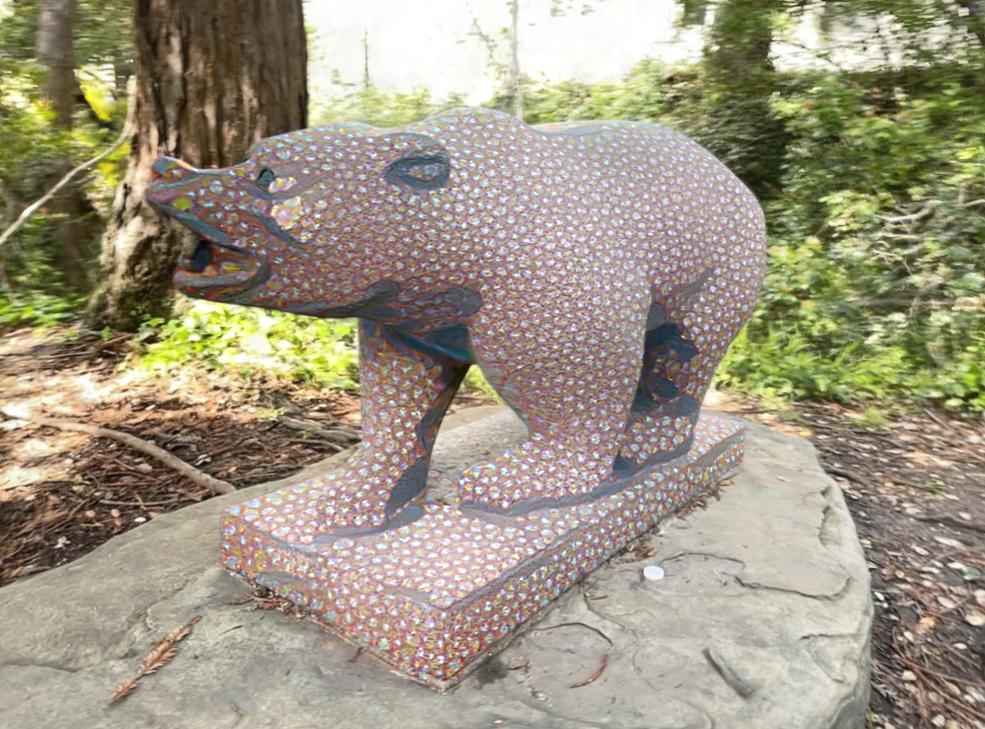} \\
        
        \includegraphics[width=2.75cm]{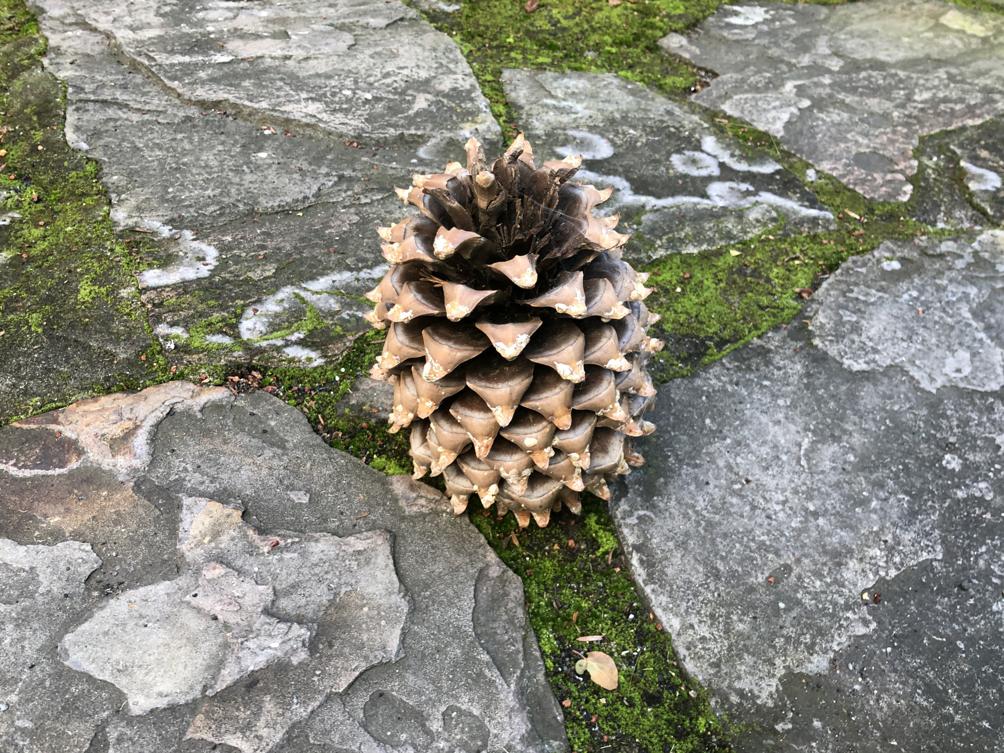} & \includegraphics[width=2.75cm]{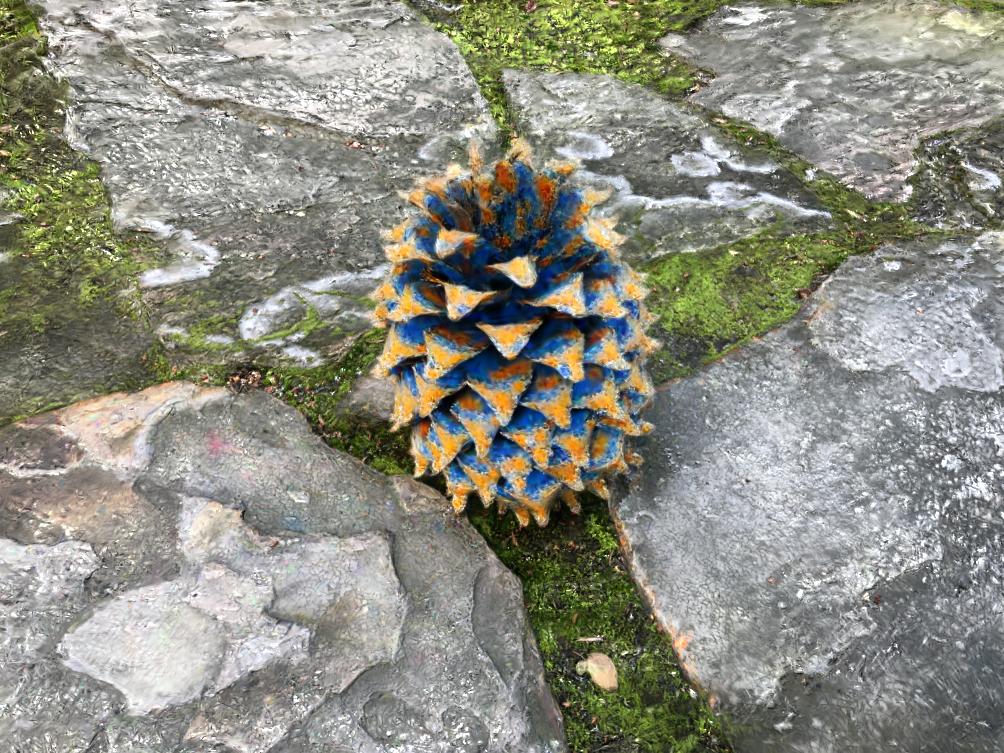} & \includegraphics[width=2.75cm]{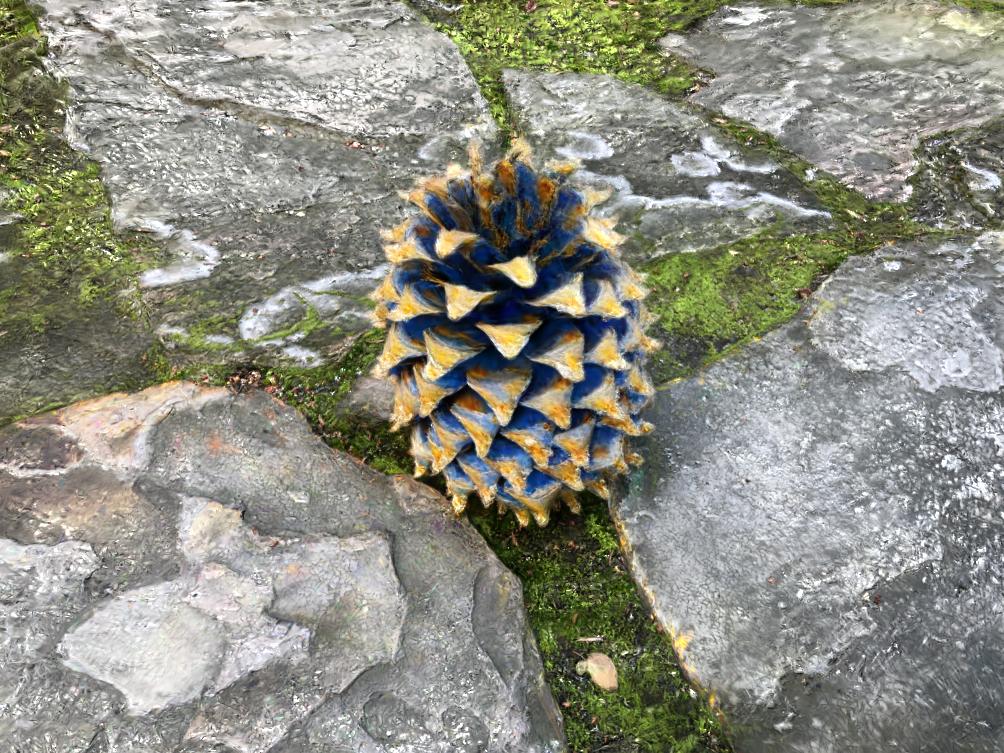} & \includegraphics[width=2.75cm]{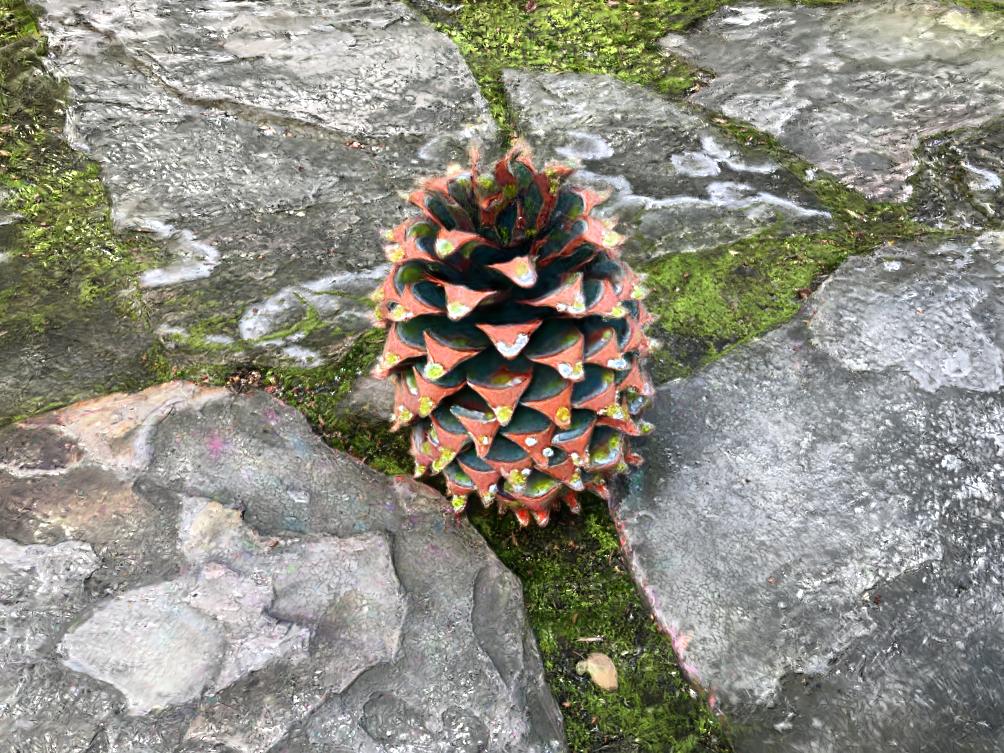} \\
         Ground Truth & Venetian Canal &  South Ledges &  Lizard Story \\
        
    \end{tabular}
    
    \caption{Shows single object style transfer on the \textit{bear} and \textit{pinecone} scenes with style images of different artistic styles and composition. Our approach localizes style transfer to the selected objects, without affecting the background. }
    \label{fig:grid}
\end{figure}

In this case, we focus on stylizing a single object from the scene. \cref{fig:grid} demonstrates that our method effectively confines the style transfer to the selected object, stylizing it according to the provided artwork. In this figure, we use three style images: Venetian Canal, South Ledges, and Lizard Story. We showcase our results on two object-centric scenes - \textit{bear}, and \textit{pinecone}. The results demonstrate precise style transfer. For example, in the \textit{bear} scene, the geometry and texture (curvature and shadows) are faithfully preserved. The adaptive nature of our style transfer, facilitated by the NNFM loss function, is evident in the \textit{pinecone} scene, where dark regions adopt a blue hue reminiscent of the Venetian Canal, while bright areas take on an orange tone.

\begin{figure}[!tb]
    \centering
    \begin{tabular}{c c c c}
        \multicolumn{4}{c}{} \\
        % 1st row: text
        & \includegraphics[width=2cm]{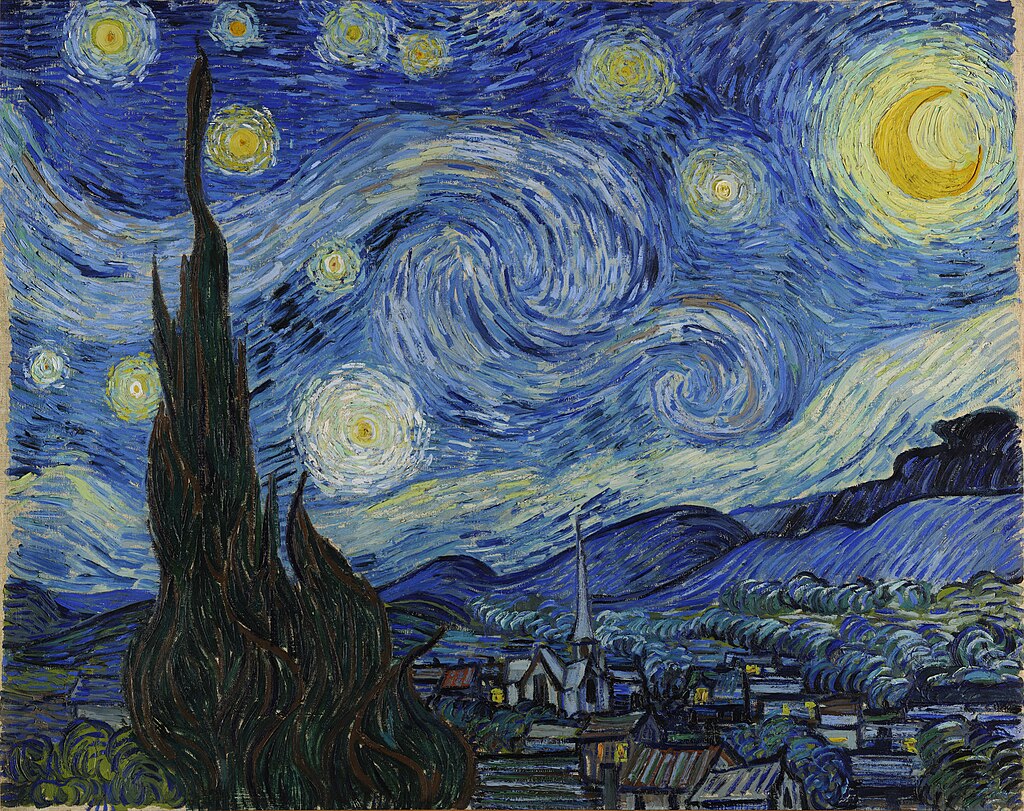} & \includegraphics[width=2cm, height=1.6cm]{new_figs/style-ims/venitian-canal.jpg} & \includegraphics[width=2cm]{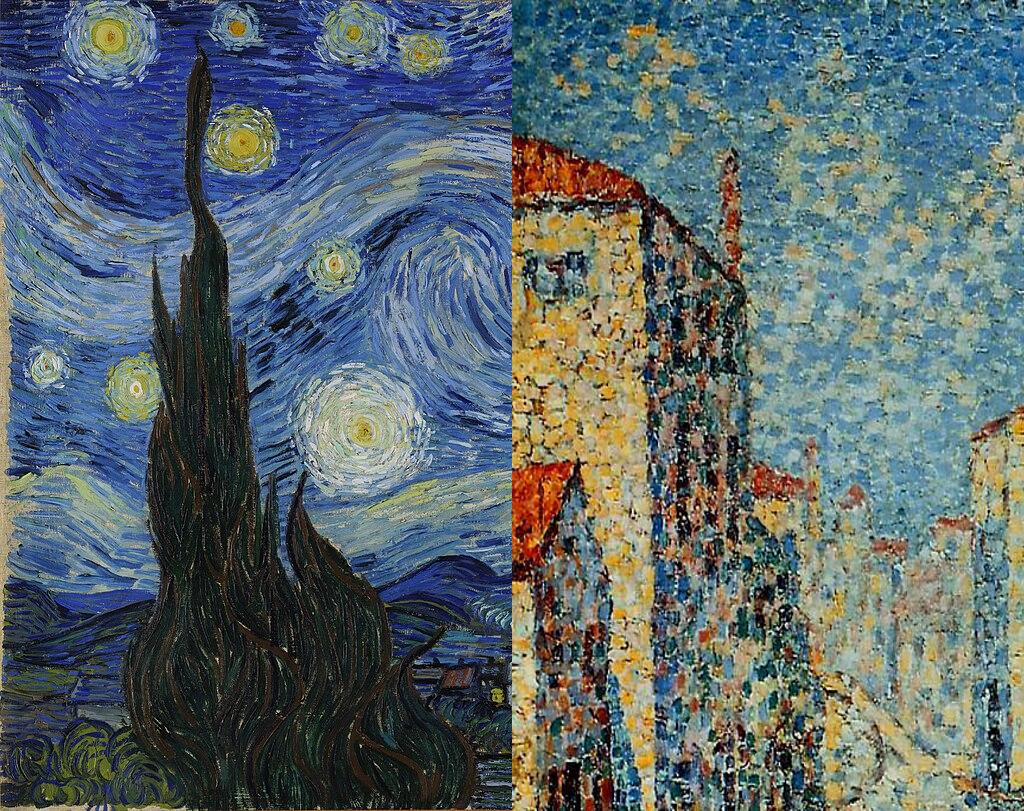} \\
         & Starry Night & Venetian Canal & Starry + Venetian  \\
        
        \includegraphics[width=2.75cm]{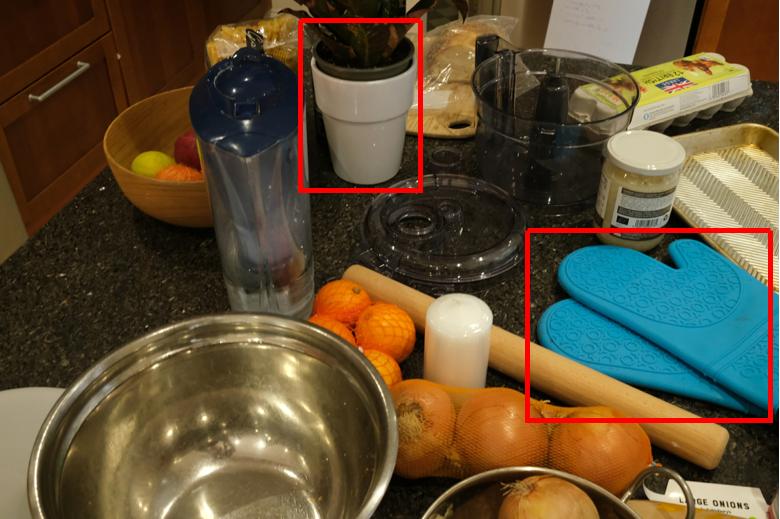} & \includegraphics[width=2.75cm]{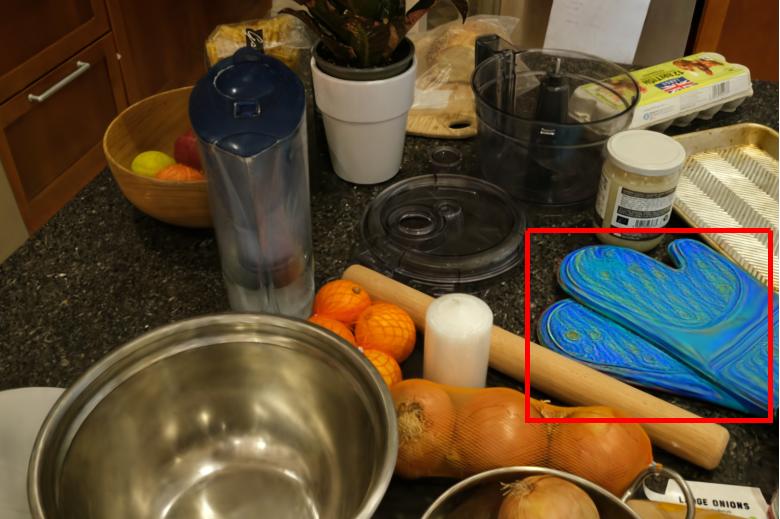} & \includegraphics[width=2.75cm]{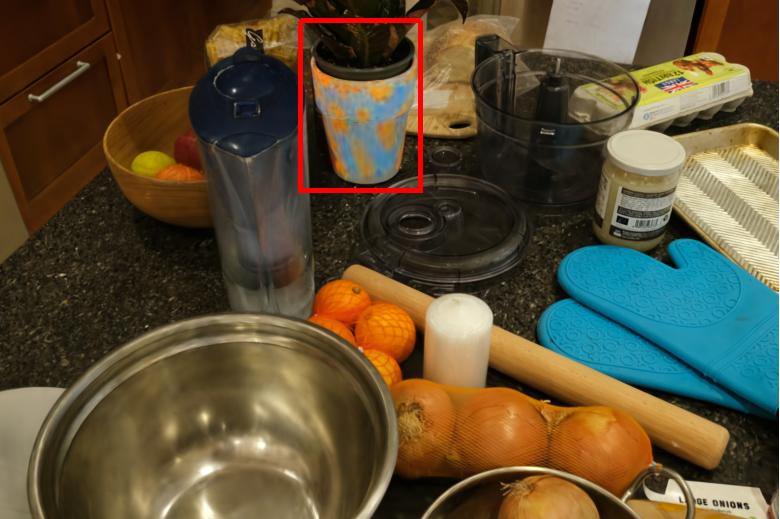} & \includegraphics[width=2.75cm]{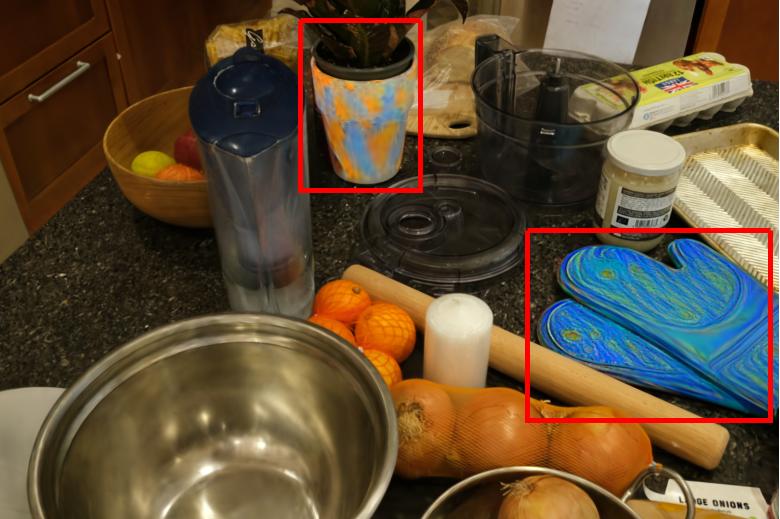} \\
        \includegraphics[width=2.75cm]{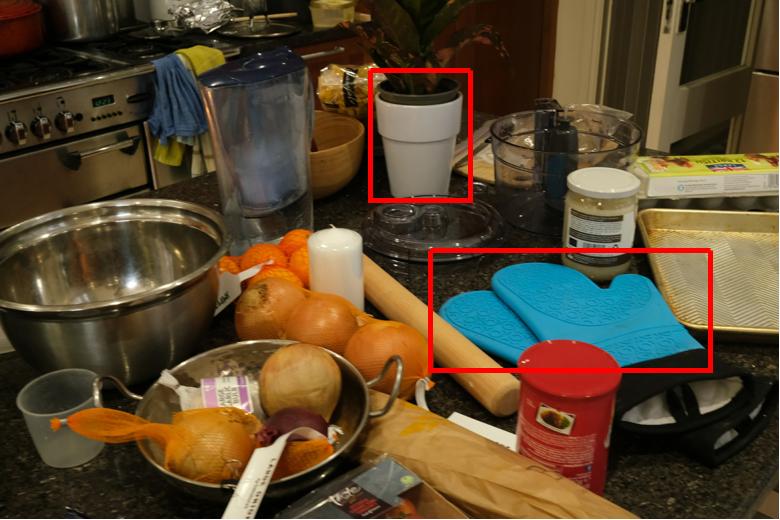} & \includegraphics[width=2.75cm]{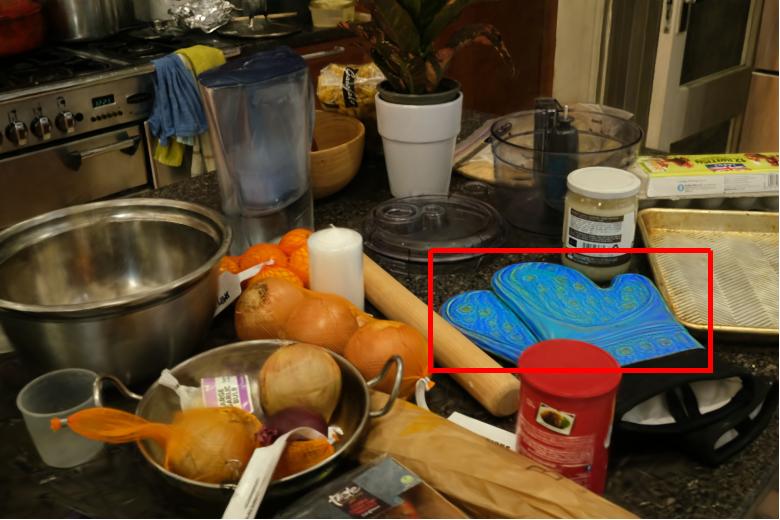} & \includegraphics[width=2.75cm]{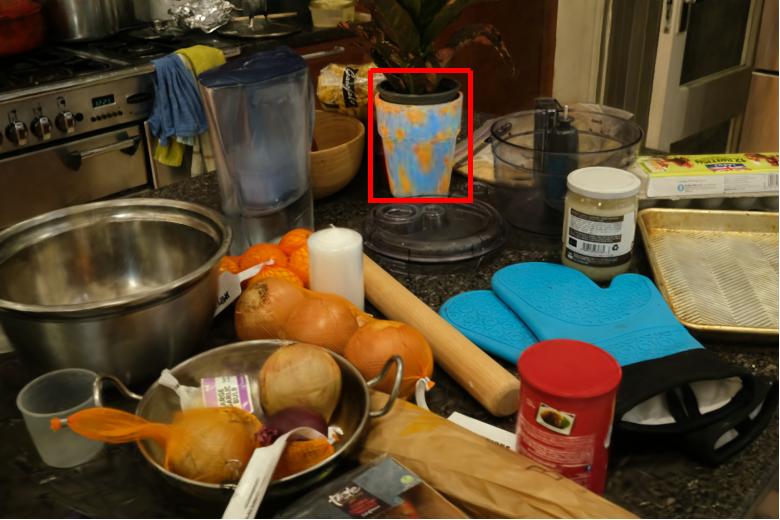} & \includegraphics[width=2.75cm]{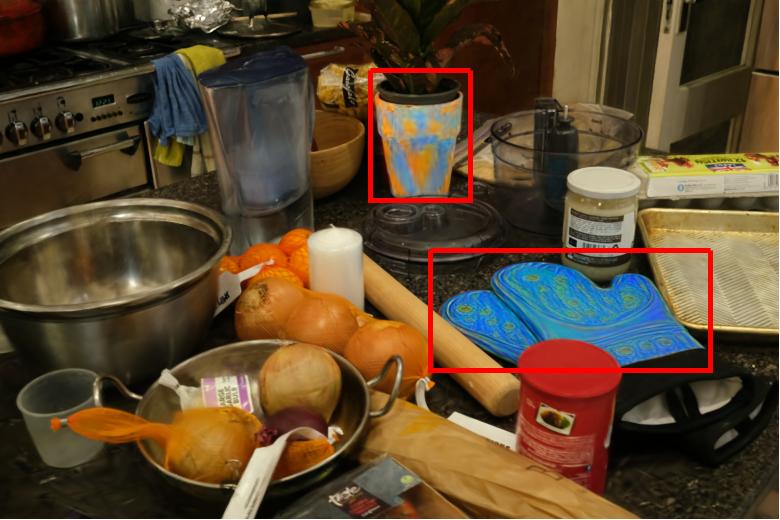} \\
        \includegraphics[width=2.75cm]{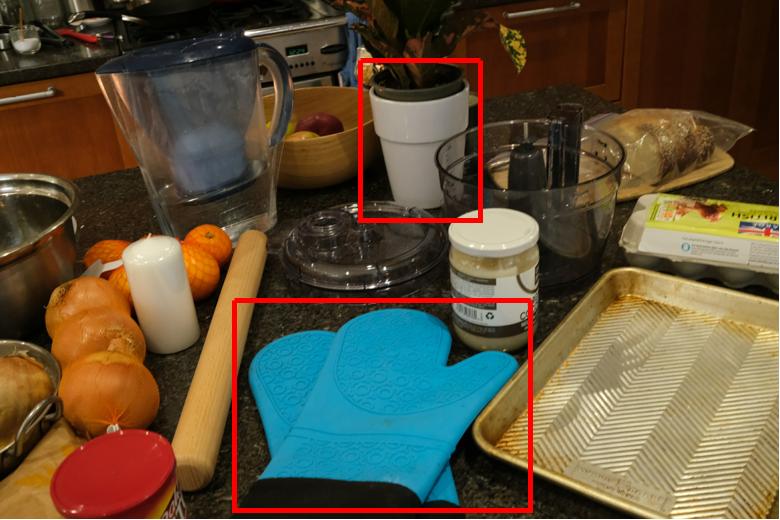} & \includegraphics[width=2.75cm]{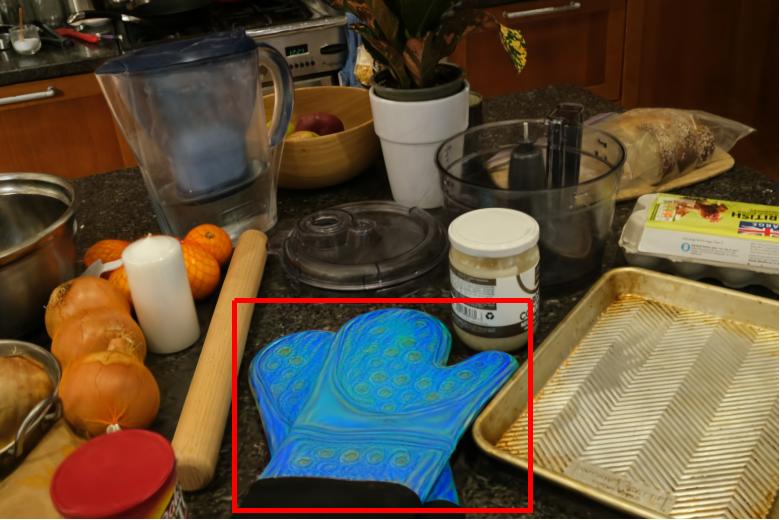} & \includegraphics[width=2.75cm]{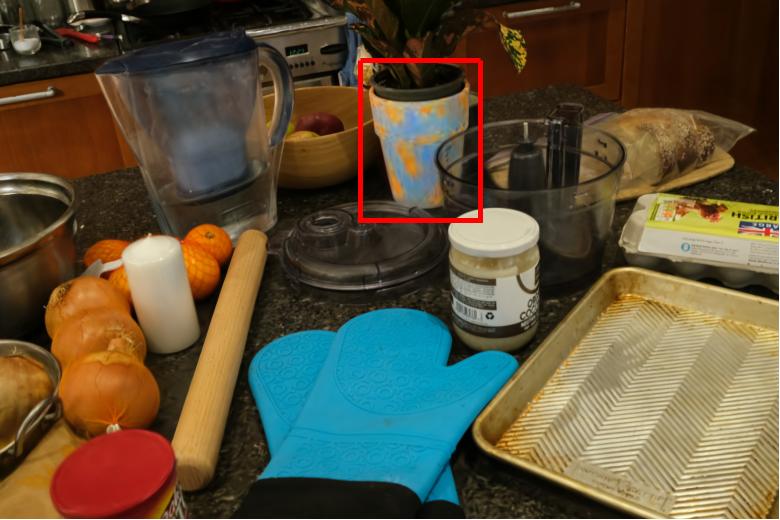} & \includegraphics[width=2.75cm]{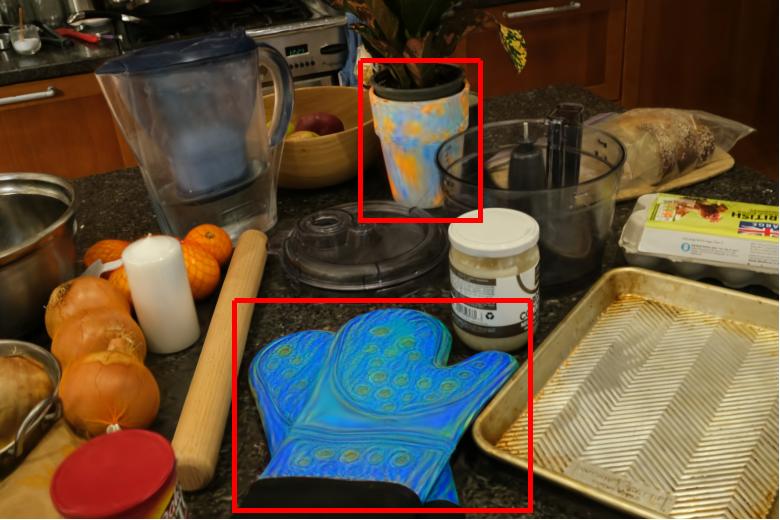} \\
        Ground Truth & Mitten Stylized  & Flower Pot Stylized & Both Stylized \\ 
        
        \includegraphics[width=2.75cm]{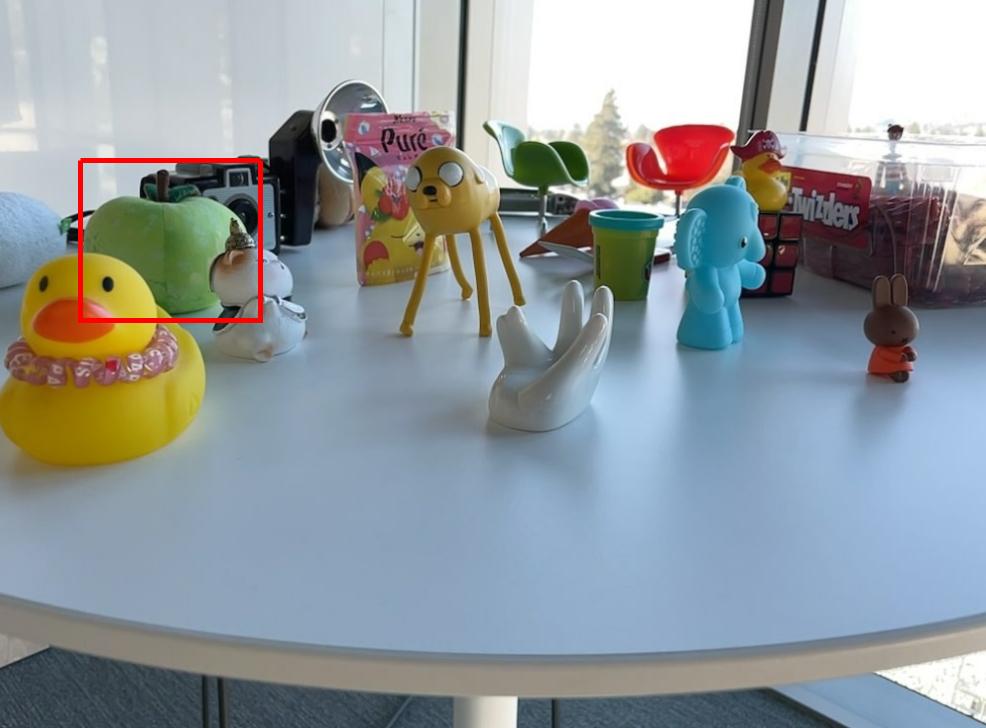} & \includegraphics[width=2.75cm]{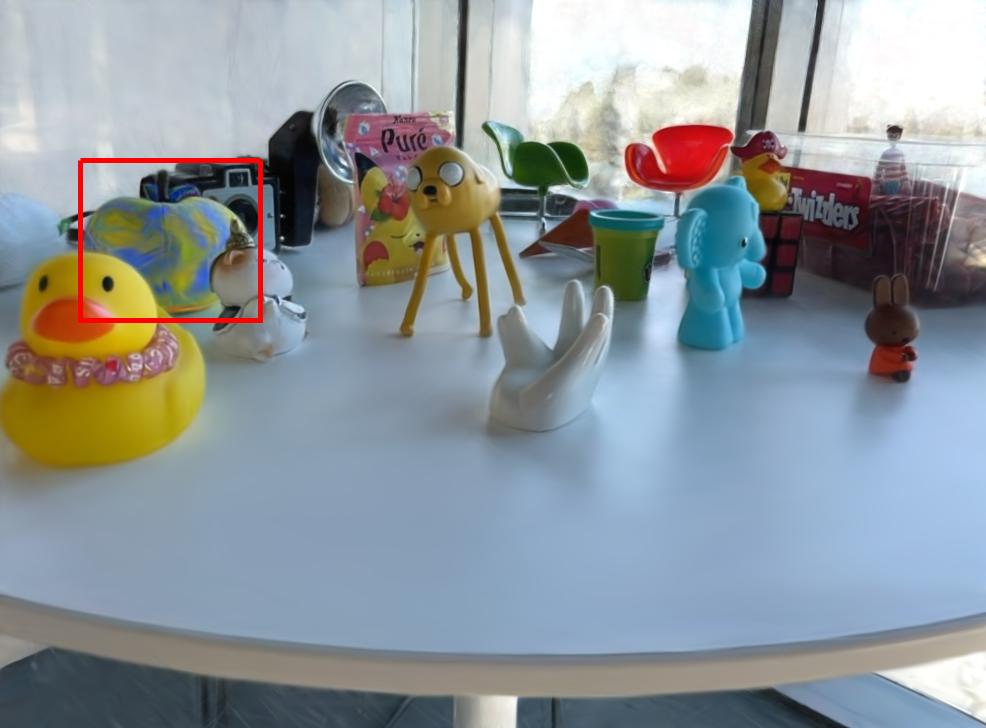} & \includegraphics[width=2.75cm]{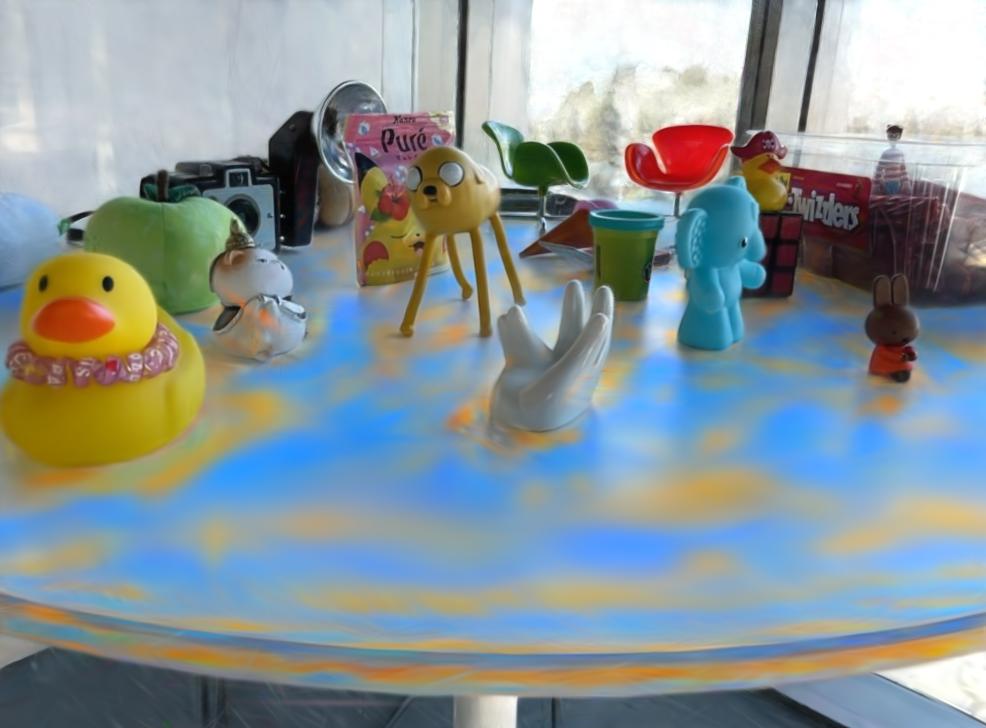} & \includegraphics[width=2.75cm]{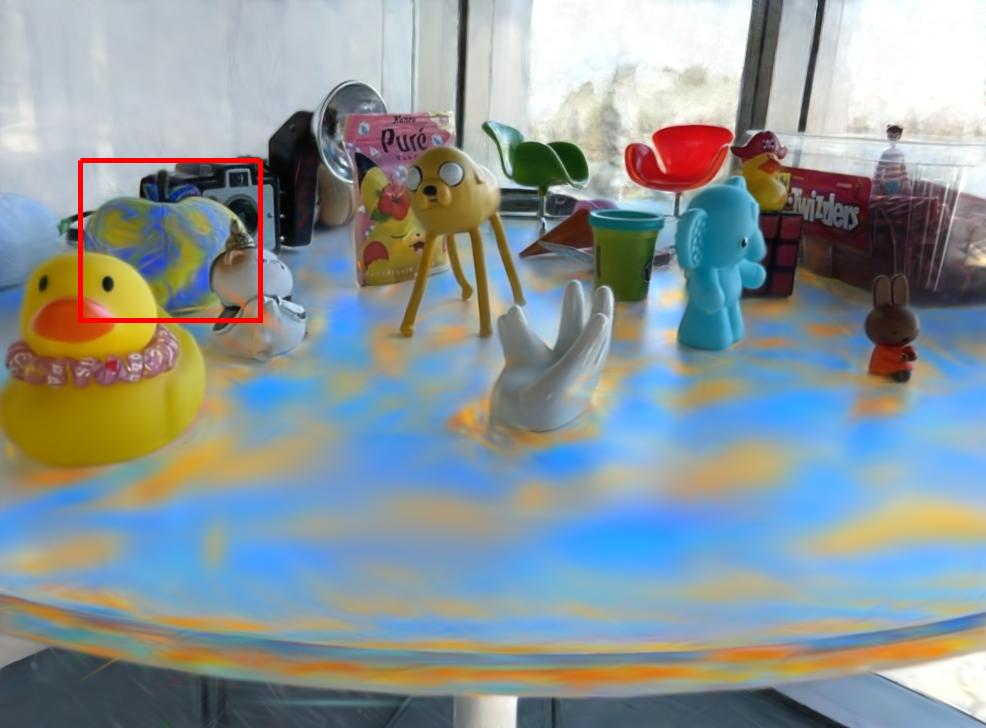} \\
        \includegraphics[width=2.75cm]{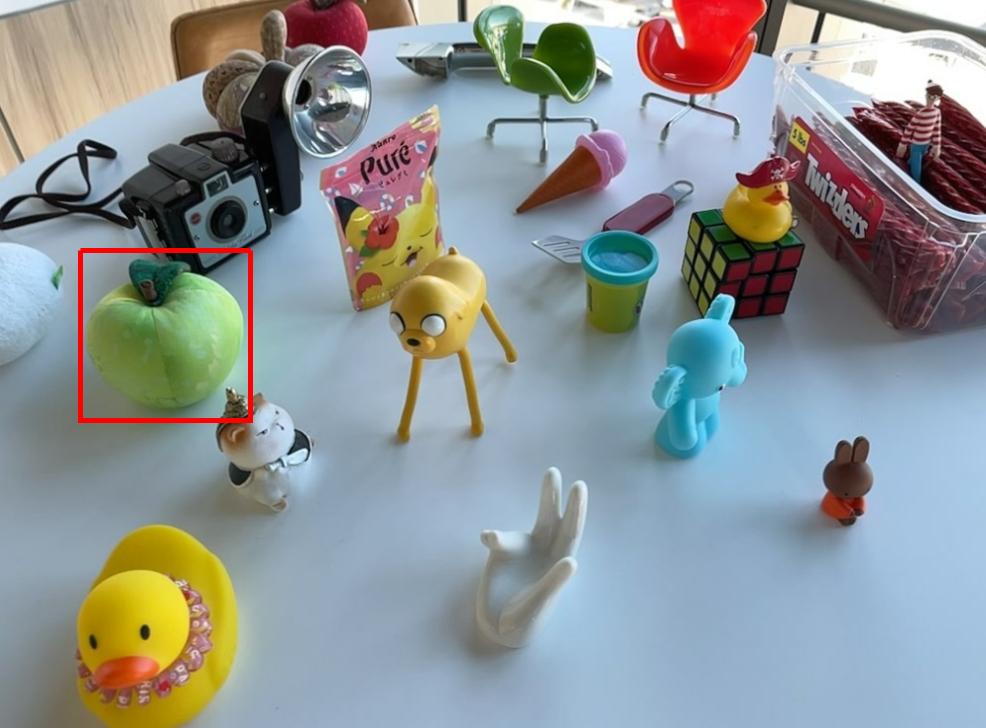} & \includegraphics[width=2.75cm]{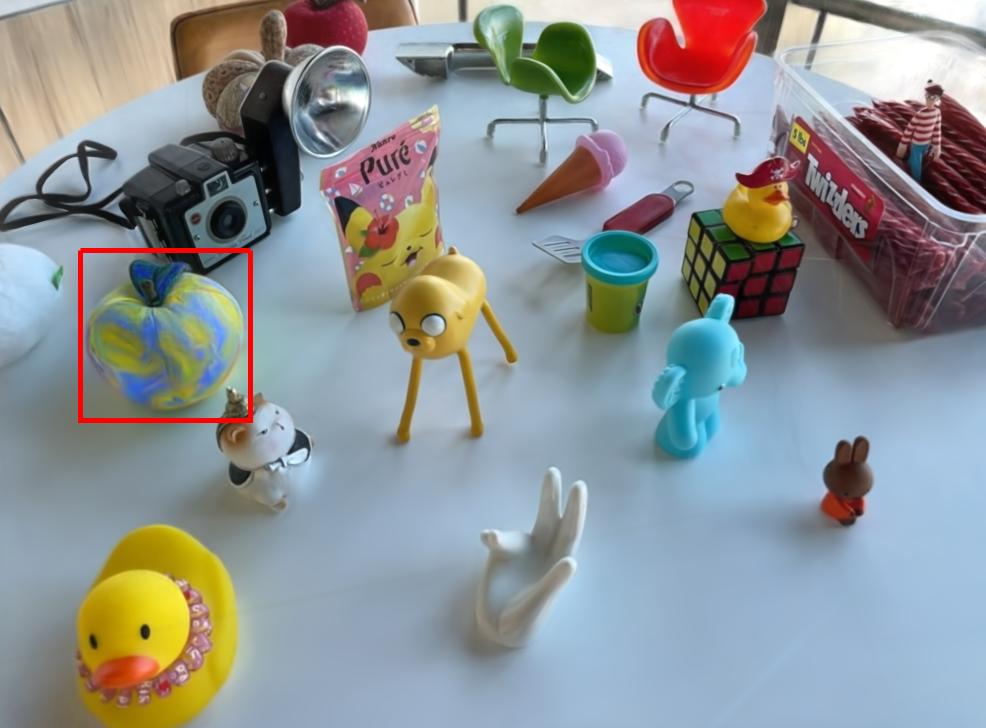} & \includegraphics[width=2.75cm]{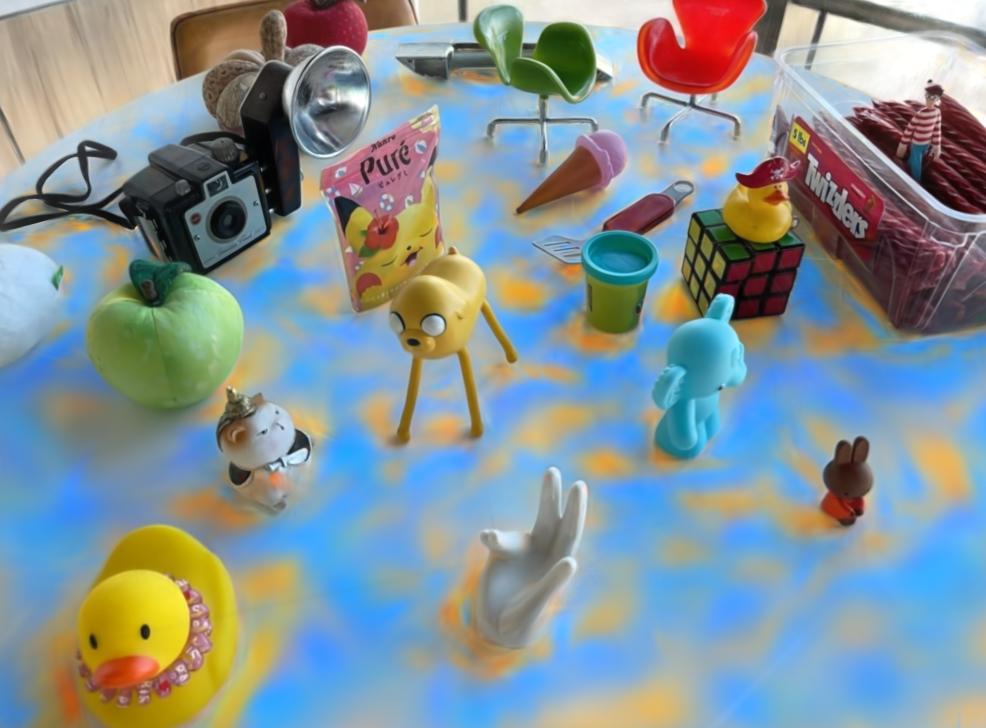} & \includegraphics[width=2.75cm]{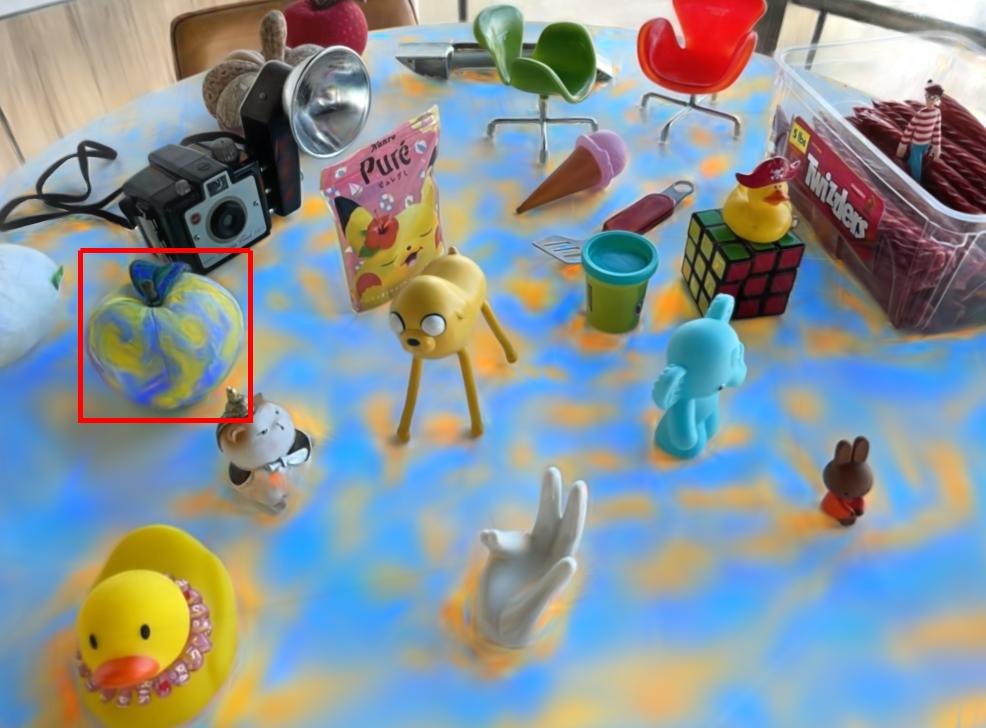} \\
        \includegraphics[width=2.75cm]{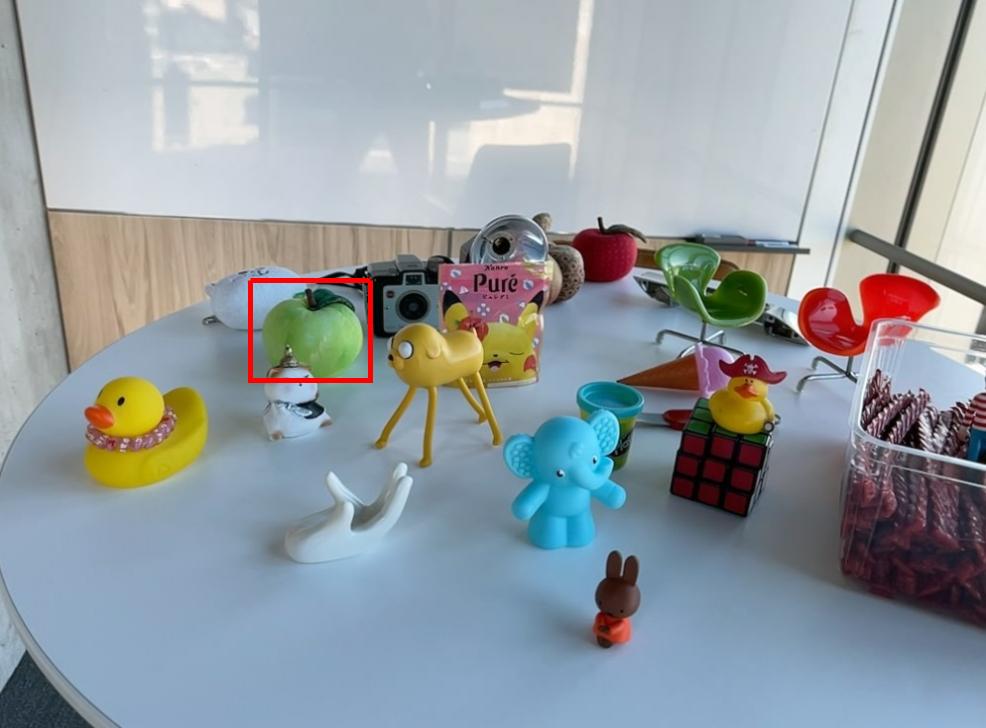} & \includegraphics[width=2.75cm]{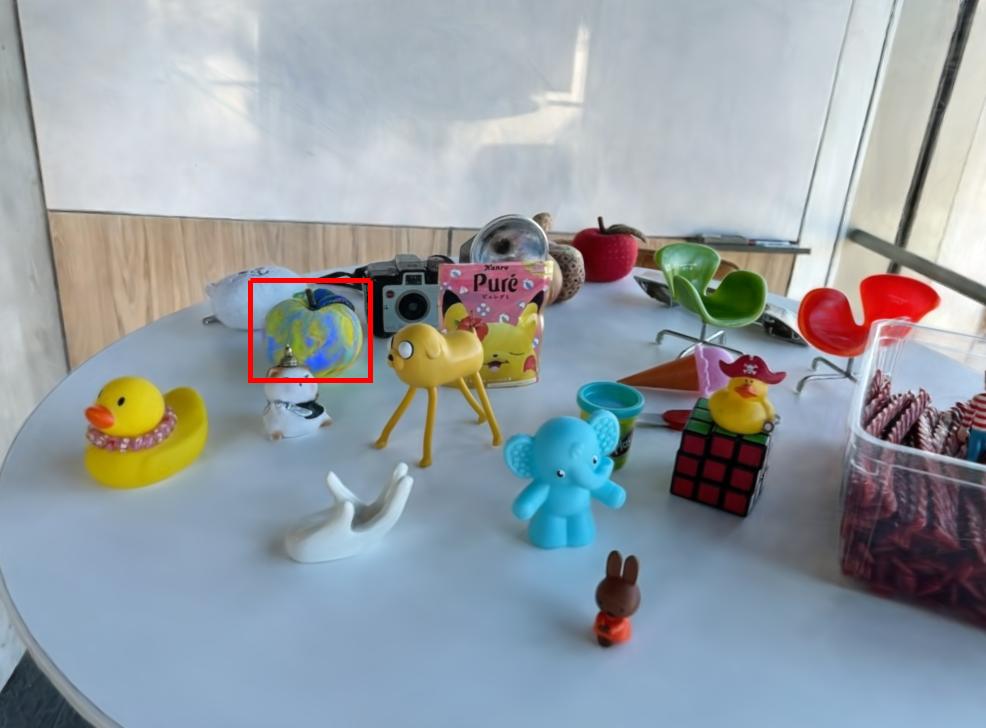} & \includegraphics[width=2.75cm]{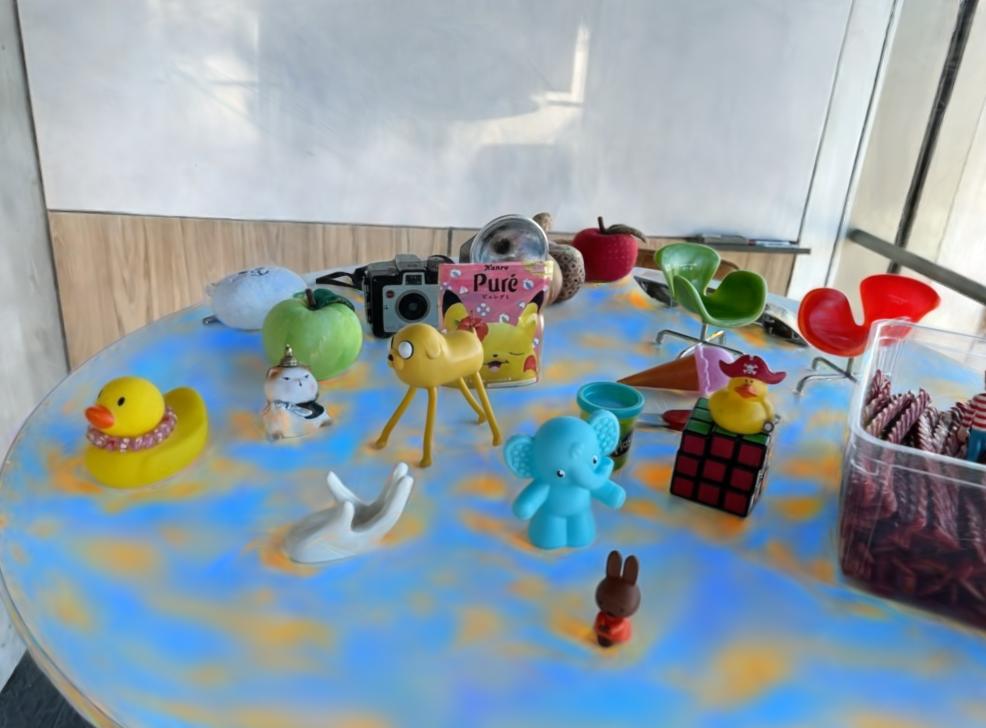} & \includegraphics[width=2.75cm]{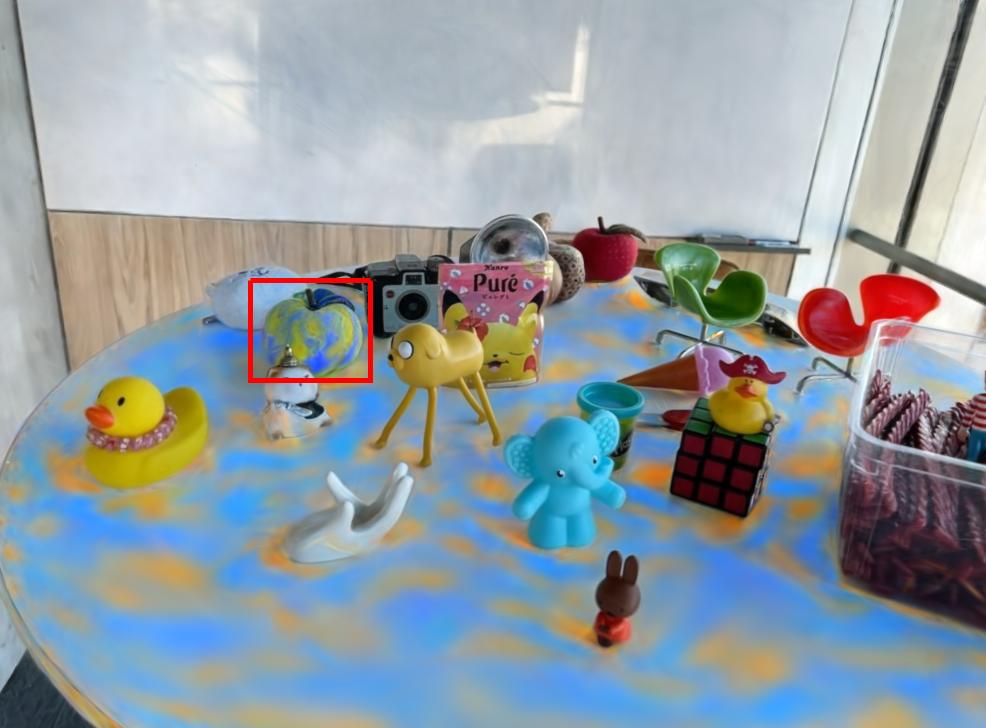} \\
        Ground Truth & Apple Stylized  & Tabletop Stylized   & Both Stylized \\ 
    \end{tabular}
    \caption{Multi-object style transfer. The style images above are applied to selected objects highlighted in the two scenes (\textit{counter} and \textit{figurines}). In columns 2 \& 3 we show stylizing of individual objects and in column 4 we see both the stylized objects together.}
    \label{fig:multi}
\end{figure}

\subsection{Multi-object style transfer} \label{multiobj} Our method extends to stylizing multiple objects within a scene, where we select two distinct objects and apply different style images to each.
\cref{fig:multi} illustrates the results for these scenes, showcasing both single-object and multi-object applications of our style transfer method across scenes featuring numerous objects.

The first half of the image shows the \textit{counter} scene from the MipNerf \cite{mipnerf} dataset. We stylize the `mitten' and the `flower pot' in this scene. From three different viewpoints, it is evident that our stylization approach is highly controllable, allowing us to accurately select and stylize objects while preserving their geometry.

In the second half of the figure, we stylize the `green apple' and the entire `tabletop' in the \textit{figurines} scene from the LERF \cite{lerflanguageembeddedradiance} dataset. Our approach is effective at stylizing the `tabletop' even though it is not in the foreground and is occluded by various objects. The stylization occurs seamlessly, making the stylized `tabletop' appear as if the style is an inherent property of the surface.

\begin{figure}[!tb]
    \centering
    \begin{tabular}{c c c}
        \multicolumn{3}{c}{} \\
        % 1st row: text
        & \includegraphics[width=2cm]{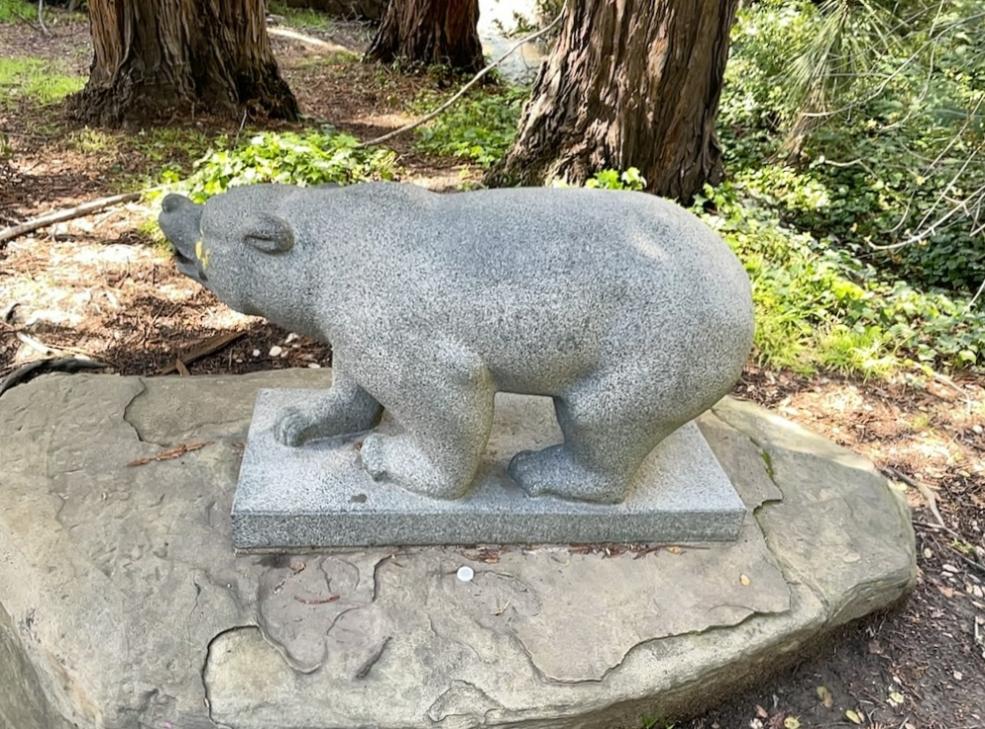} & \includegraphics[width=2cm, height=1.6cm]{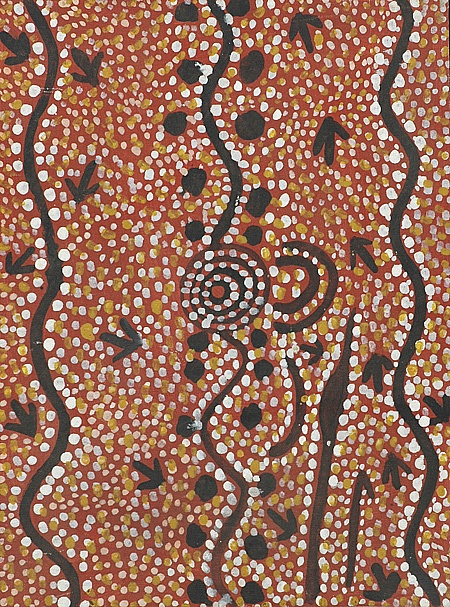} \\
         & Ground Truth & Style Image  \\
        
        \includegraphics[width=2.75cm]{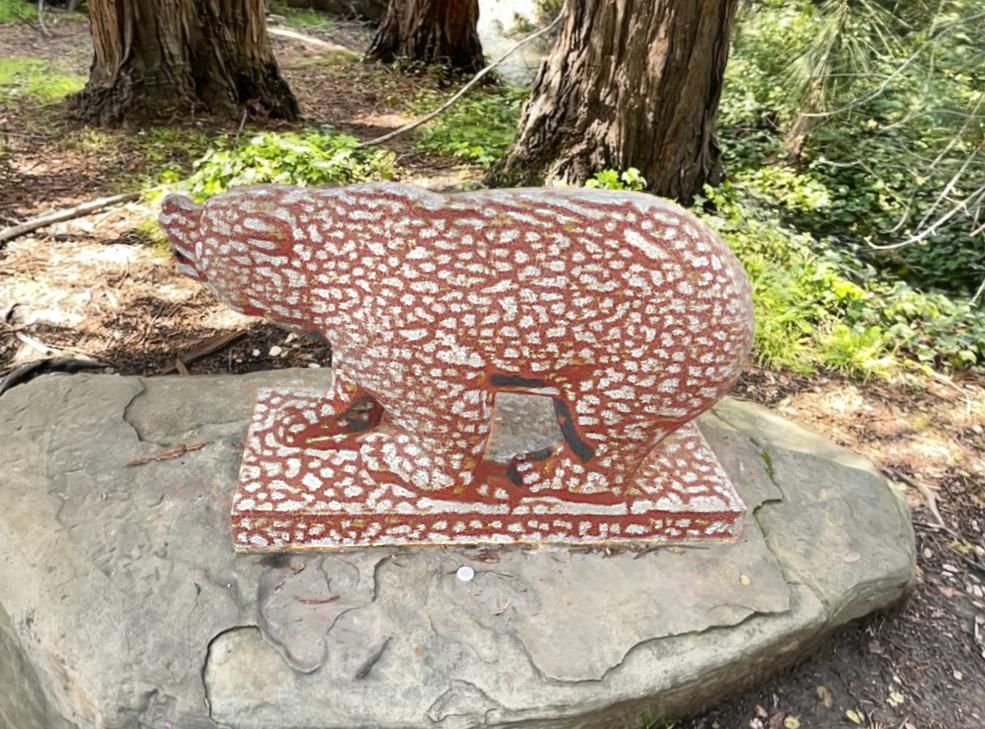} & \includegraphics[width=2.75cm]{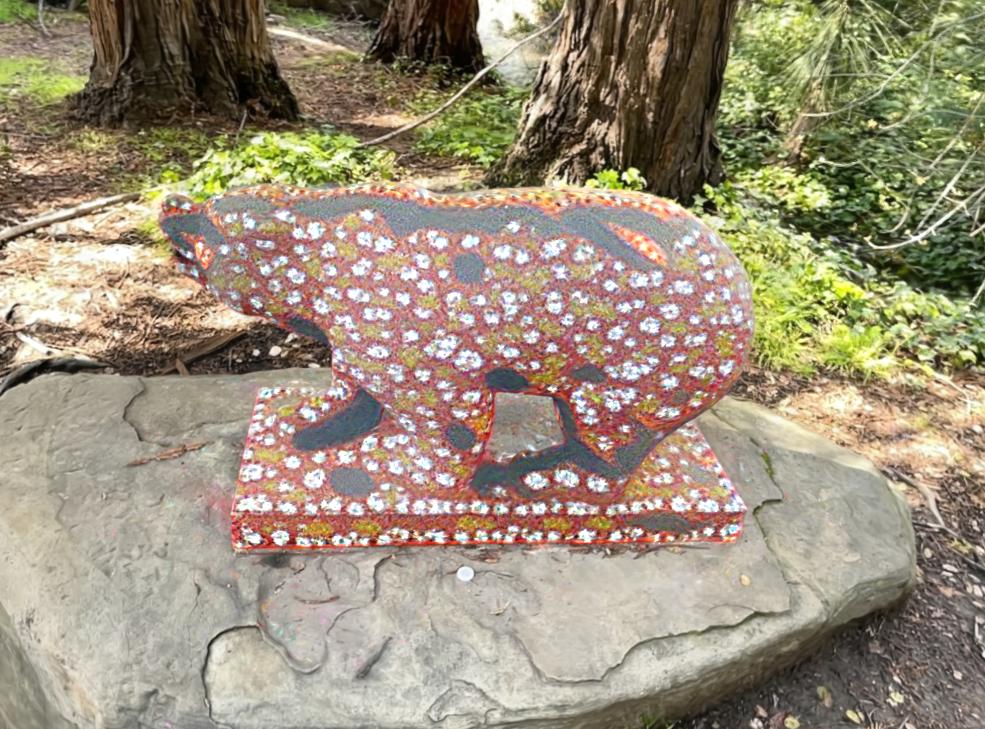} & \includegraphics[width=2.75cm]{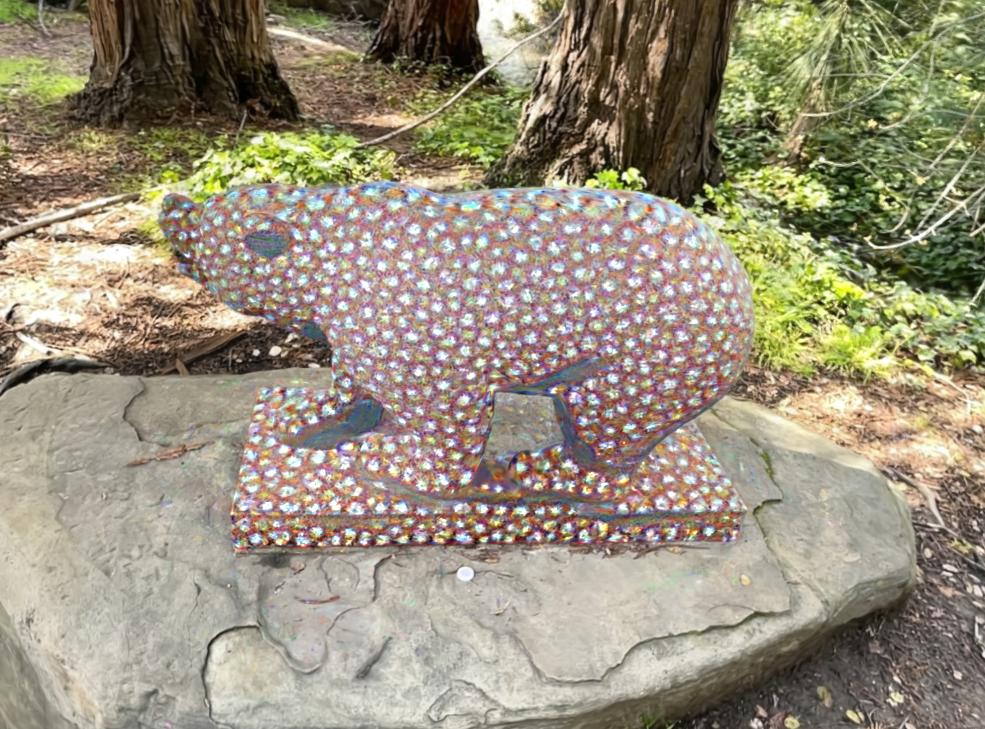} \\
        Layers: [1, 3] & Layers: [6, 8]  & Layers: [11, 13, 15] \\
        
        \includegraphics[width=2.75cm]{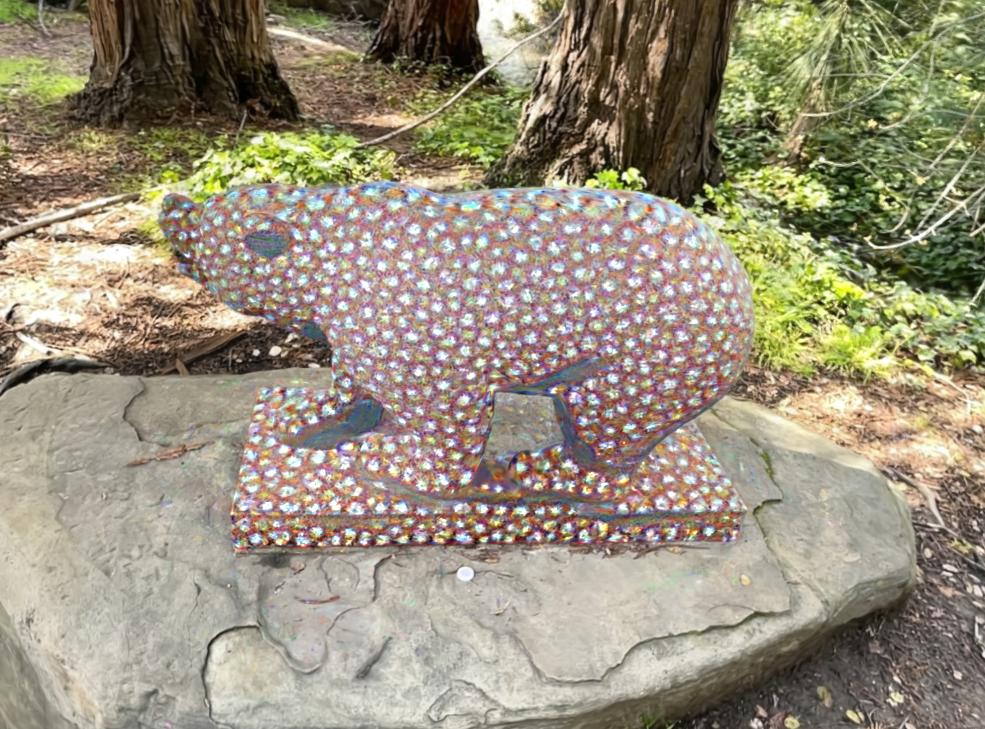} & \includegraphics[width=2.75cm]{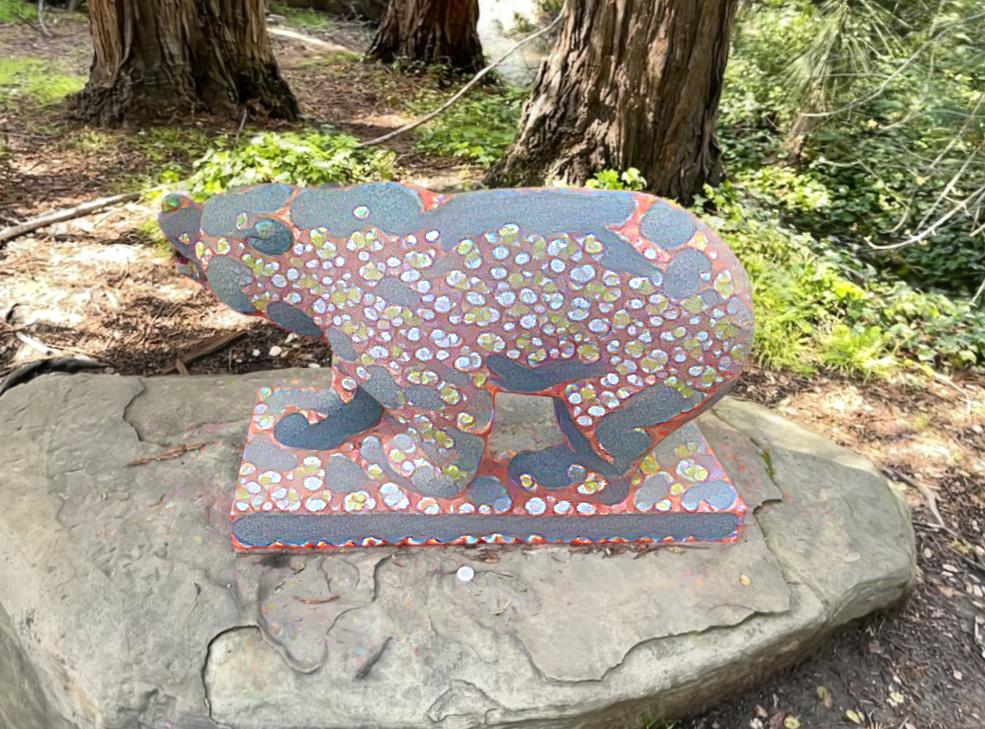} & \includegraphics[width=2.75cm]{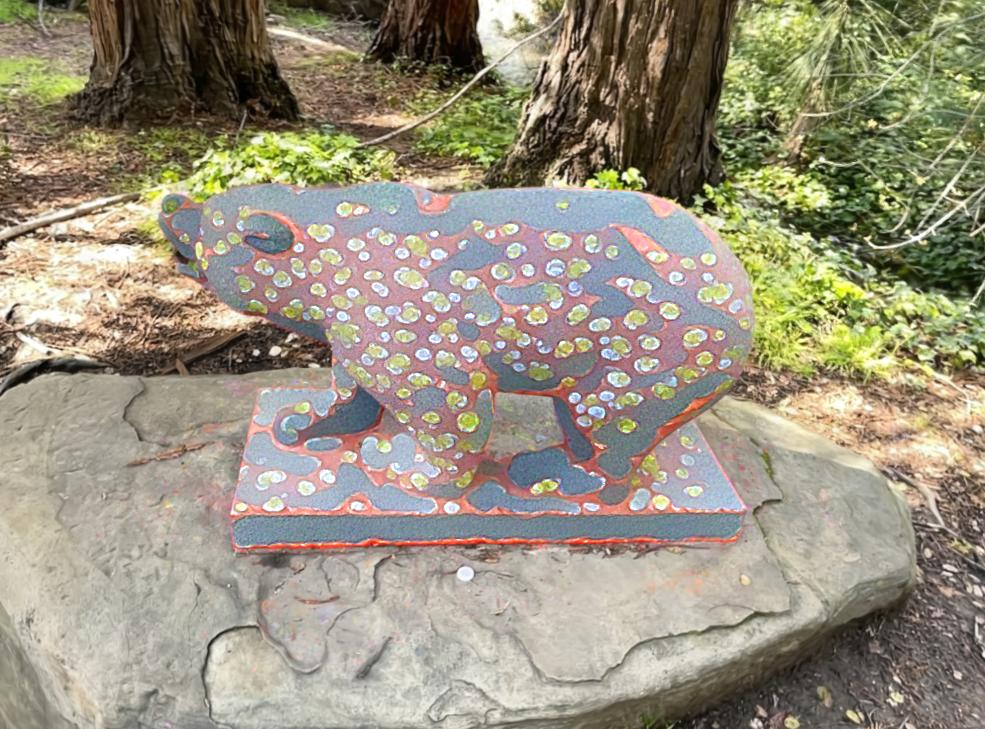} \\
        Scale: 0.25 & Scale: 0.5 & Scale: 0.75 \\ 

        \includegraphics[width=2.75cm]{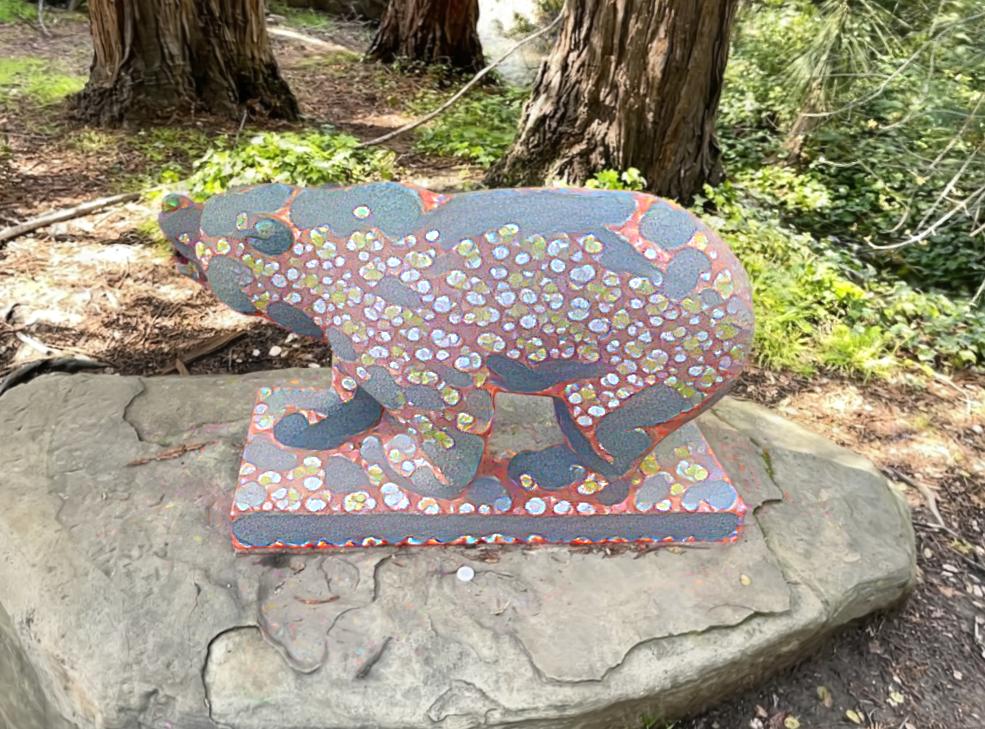} & \includegraphics[width=2.75cm]{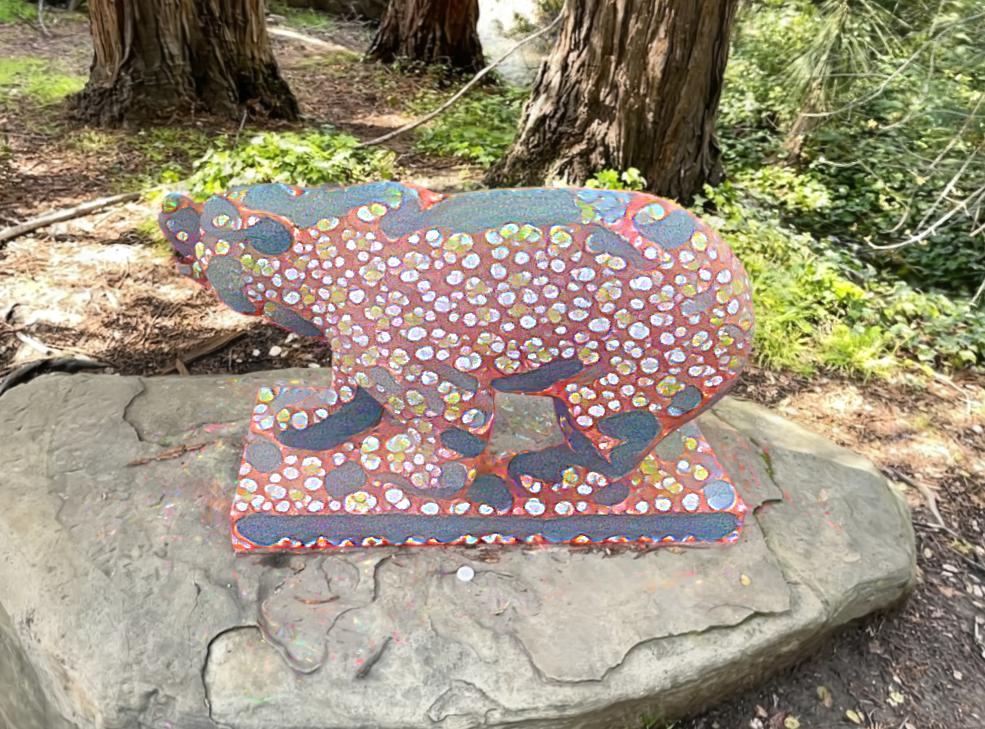} & \includegraphics[width=2.75cm]{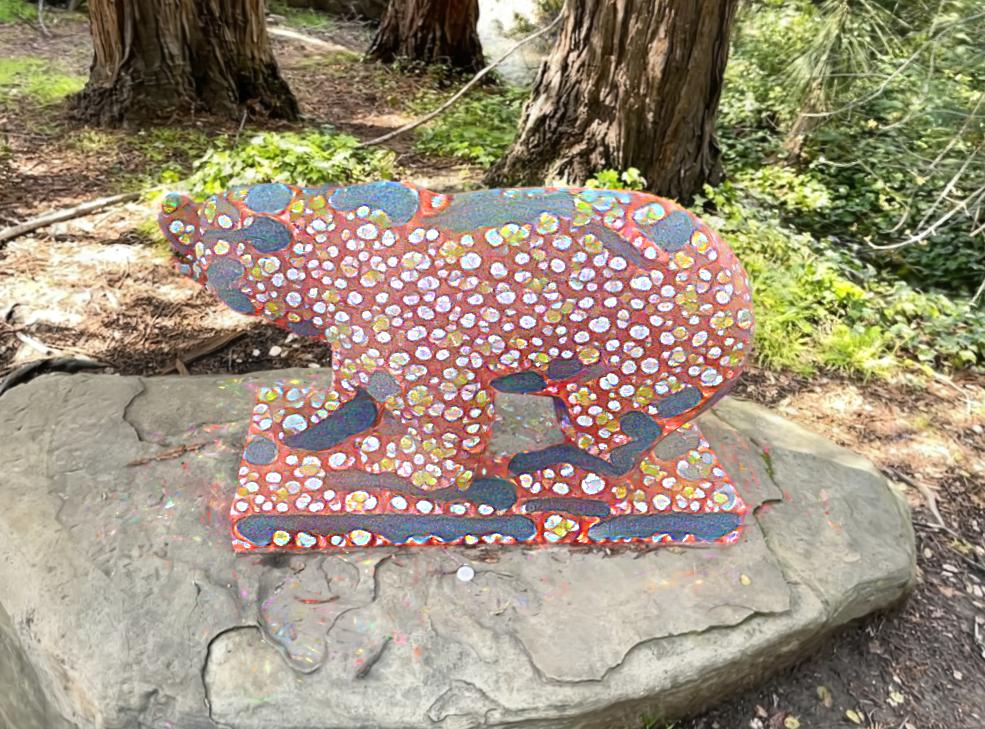} \\
        LR: 0.025 & LR: 0.05 & LR: 0.075 \\ 
        
    \end{tabular}
    \caption{Visualisation of scale and intensity control. The scale of the style pattern can be controlled by changing the size of the style image or changing the layers for feature extraction. Intensity can be controlled by varying the learning rate. 
    % We use the image of a bear as an object-centric scene, and stylize a patter of a \textit{lizard} on it. And we show the results of  stylization when \textbf{a)} different layers of the VGG backbone are used to extract features in order to calculate the nNFM and \textbf{b)} when different resolutions of the style image are used tweaking its scaling factor.
    }
    \label{fig:ablation_nnfm}
\end{figure}

\subsection{Scale Control} \label{subsec:control}
Previous works \cite{jing2018strokecontrollablefaststyle, li2023arfpluscontrollingperceptualfactors} have shown that the scale of the style patterns can be controlled by two parameters - the receptive field of the VGG features and the size of the style image. The receptive field can be controlled by varying the layers of the VGG network which are used to extract features for the NNFM loss. The size of the style image can be controlled by a downscaling factor (scale = 1 means that the style image is used in its original resolution). 

In \cref{fig:ablation_nnfm}, we present experimental results exploring these scale parameters and their impact on stylization. We can observe that as the scale of the style image is decreased, the repetition of the pattern increases. A similar effect can be observed with the network layer selection. The features from the deeper layers have larger receptive fields, giving an effect similar to downscaling the style image. It is interesting to note that if we use layers [1,3] which are somewhat early in the network, the features extracted from those layers for style transfer are not able to accurately learn patterns in the style image.

Additionally, we show that the intensity of style transfer can be controlled by the learning rate, with a larger rate giving stronger style transfer.

\begin{figure}[!tb]
    \centering
    \includegraphics[width=0.8\linewidth]{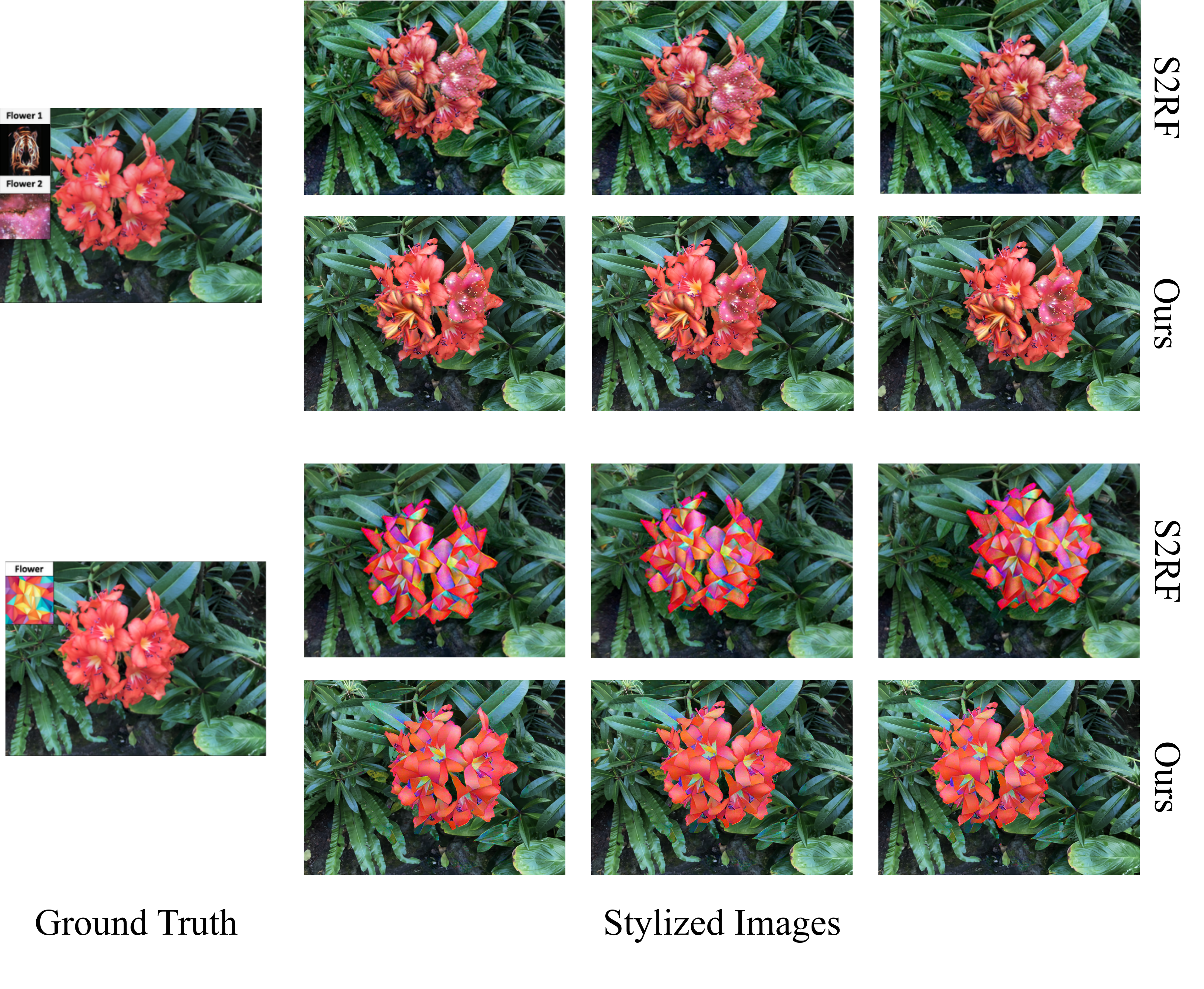}
    \caption{Qualitative comparison of our method with S2RF on the \textit{flower} scene.}
    \label{fig:comparison}
\end{figure}

\subsection{Qualitative Comparison} \label{subsec:s2rf}
We evaluate our approach in comparison to S2RF\cite{lahiri2023s2rf}, which also focuses on localized stylization of specific objects rather than entire 3D scenes. However, S2RF introduces discoloration in non-target areas, likely due to its use of a grid-based representation which shares parameters across local regions. This parameter sharing can inadvertently affect neighboring regions, which is undesirable for this task. In contrast, our method utilizes segmented 3D Gaussians to precisely isolate style transfer to the target object, leaving other Gaussians unaffected. This distinction is evident in the first two rows in \cref{fig:comparison}, which illustrates the differences between the two approaches when transferring style on two specific flowers. Our method also performs better in preserving content of the ground truth images. In the last two rows, we observe that in the stylized image generated by S2RF, the shape of the individual flowers is difficult to discern, whereas our approach achieves a more realistic style transfer result.

Additionally, S2RF achieves an average rendering speed of 15 FPS whereas our approach achieves 100+ FPS due to its reliance on the 3DGS representation.

\section{Limitations}
While our method provides efficient and controllable stylization of 3D Gaussian splats, it has certain limitations. Firstly, geometric artifacts arising from the initial 3DGS reconstruction process can occasionally affect the quality of the final stylized scenes. Additionally, the use of the Segment Anything Model for view-specific segmentation can sometimes struggle with generating detailed masks from particular angles, leading to the unintended merging of object parts. Consequently, edits intended for specific areas might inadvertently impact adjacent regions. Although these issues are rare in most scenarios, it can occur in complex scenes or areas with poorly defined boundaries.

\section{Conclusion}
In this work, we introduce StyleSplat, a lightweight technique to stylize 3D objects using a reference style image in the 3D Gaussian Splatting paradigm. StyleSplat leverages off-the-shelf image segmentation and tracking models to obtain consistent 2D masks. These masks are then used as supervision for segmenting 3D Gaussians into distinct objects while jointly optimizing their geometry and color. Finally, a nearest-neighbor feature matching loss is used to finetune the selected Gaussians by aligning their spherical harmonic coefficients with the provided style image. After the initial training and segmentation, StyleSplat takes less than a minute to perform stylization, allowing fast experimentation. We showcase the effectiveness of our method on a variety of scenes and styles, highlighting its suitability for artistic endeavors.

\bibliographystyle{splncs04}
\bibliography{main}

\begin{thebibliography}{10}
\providecommand{\url}[1]{\texttt{#1}}
\providecommand{\urlprefix}{URL }
\providecommand{\doi}[1]{https://doi.org/#1}

\bibitem{wikiartWikiArtorgVisual}
{W}iki{A}rt.org - {V}isual {A}rt {E}ncyclopedia --- wikiart.org. \url{https://www.wikiart.org/}, [Accessed 30-06-2024]

\bibitem{mipnerf}
Barron, J.T., Mildenhall, B., Verbin, D., Srinivasan, P.P., Hedman, P.: Mip-nerf 360: Unbounded anti-aliased neural radiance fields. CoRR  \textbf{abs/2111.12077} (2021), \url{https://arxiv.org/abs/2111.12077}

\bibitem{dino}
Caron, M., Touvron, H., Misra, I., Jégou, H., Mairal, J., Bojanowski, P., Joulin, A.: Emerging properties in self-supervised vision transformers (2021), \url{https://arxiv.org/abs/2104.14294}

\bibitem{castillo2017sonzornslemmatargeted}
Castillo, C., De, S., Han, X., Singh, B., Yadav, A.K., Goldstein, T.: Son of zorn's lemma: Targeted style transfer using instance-aware semantic segmentation (2017), \url{https://arxiv.org/abs/1701.02357}

\bibitem{GaussianEditor}
Chen, Y., Chen, Z., Zhang, C., Wang, F., Yang, X., Wang, Y., Cai, Z., Yang, L., Liu, H., Lin, G.: Gaussianeditor: Swift and controllable 3d editing with gaussian splatting. In: Proceedings of the IEEE/CVF Conference on Computer Vision and Pattern Recognition (CVPR). pp. 21476--21485 (June 2024)

\bibitem{cheng2023tracking}
Cheng, H.K., Oh, S.W., Price, B., Schwing, A., Lee, J.Y.: Tracking anything with decoupled video segmentation (2023)

\bibitem{gatys2015neuralalgorithmartisticstyle}
Gatys, L.A., Ecker, A.S., Bethge, M.: A neural algorithm of artistic style (2015), \url{https://arxiv.org/abs/1508.06576}

\bibitem{neuralstyletransfer}
Gatys, L.A., Ecker, A.S., Bethge, M.: Image style transfer using convolutional neural networks. 2016 IEEE Conference on Computer Vision and Pattern Recognition (CVPR) pp. 2414--2423 (2016), \url{https://api.semanticscholar.org/CorpusID:206593710}

\bibitem{gu2018arbitrarystyletransferdeep}
Gu, S., Chen, C., Liao, J., Yuan, L.: Arbitrary style transfer with deep feature reshuffle (2018), \url{https://arxiv.org/abs/1805.04103}

\bibitem{huang2017arbitrarystyletransferrealtime}
Huang, X., Belongie, S.: Arbitrary style transfer in real-time with adaptive instance normalization (2017), \url{https://arxiv.org/abs/1703.06868}

\bibitem{iceg}
Jaganathan, V., Huang, H.H., Irshad, M.Z., Jampani, V., Raj, A., Kira, Z.: Ice-g: Image conditional editing of 3d gaussian splats (2024)

\bibitem{jing2018strokecontrollablefaststyle}
Jing, Y., Liu, Y., Yang, Y., Feng, Z., Yu, Y., Tao, D., Song, M.: Stroke controllable fast style transfer with adaptive receptive fields (2018), \url{https://arxiv.org/abs/1802.07101}

\bibitem{johnson2016perceptuallossesrealtimestyle}
Johnson, J., Alahi, A., Fei-Fei, L.: Perceptual losses for real-time style transfer and super-resolution (2016), \url{https://arxiv.org/abs/1603.08155}

\bibitem{Kerbl20233DGS}
Kerbl, B., Kopanas, G., Leimkuehler, T., Drettakis, G.: 3d gaussian splatting for real-time radiance field rendering. ACM Transactions on Graphics (TOG)  \textbf{42},  1 -- 14 (2023), \url{https://api.semanticscholar.org/CorpusID:259267917}

\bibitem{lerflanguageembeddedradiance}
Kerr, J., Kim, C.M., Goldberg, K., Kanazawa, A., Tancik, M.: Lerf: Language embedded radiance fields (2023), \url{https://arxiv.org/abs/2303.09553}

\bibitem{kirillov2023segment}
Kirillov, A., Mintun, E., Ravi, N., Mao, H., Rolland, C., Gustafson, L., Xiao, T., Whitehead, S., Berg, A.C., Lo, W.Y., Dollár, P., Girshick, R.: Segment anything (2023)

\bibitem{kolkin2019styletransferrelaxedoptimal}
Kolkin, N., Salavon, J., Shakhnarovich, G.: Style transfer by relaxed optimal transport and self-similarity (2019), \url{https://arxiv.org/abs/1904.12785}

\bibitem{kurzman2019classbasedstylingrealtimelocalized}
Kurzman, L., Vazquez, D., Laradji, I.: Class-based styling: Real-time localized style transfer with semantic segmentation (2019), \url{https://arxiv.org/abs/1908.11525}

\bibitem{lahiri2023s2rf}
Lahiri, D., Panse, N., Kumar, M.: S2rf: Semantically stylized radiance fields (2023)

\bibitem{lseg}
Li, B., Weinberger, K.Q., Belongie, S., Koltun, V., Ranftl, R.: Language-driven semantic segmentation (2022), \url{https://arxiv.org/abs/2201.03546}

\bibitem{li2023arfpluscontrollingperceptualfactors}
Li, W., Wu, T., Zhong, F., Oztireli, C.: Arf-plus: Controlling perceptual factors in artistic radiance fields for 3d scene stylization (2023), \url{https://arxiv.org/abs/2308.12452}

\bibitem{liao2017visualattributetransferdeep}
Liao, J., Yao, Y., Yuan, L., Hua, G., Kang, S.B.: Visual attribute transfer through deep image analogy (2017), \url{https://arxiv.org/abs/1705.01088}

\bibitem{liu2024stylegaussian}
Liu, K., Zhan, F., Xu, M., Theobalt, C., Shao, L., Lu, S.: Stylegaussian: Instant 3d style transfer with gaussian splatting (2024)

\bibitem{liu2023grounding}
Liu, S., Zeng, Z., Ren, T., Li, F., Zhang, H., Yang, J., Li, C., Yang, J., Su, H., Zhu, J., et~al.: Grounding dino: Marrying dino with grounded pre-training for open-set object detection. arXiv preprint arXiv:2303.05499  (2023)

\bibitem{mildenhall2020nerfrepresentingscenesneural}
Mildenhall, B., Srinivasan, P.P., Tancik, M., Barron, J.T., Ramamoorthi, R., Ng, R.: Nerf: Representing scenes as neural radiance fields for view synthesis (2020), \url{https://arxiv.org/abs/2003.08934}

\bibitem{mueller2022instant}
M\"uller, T., Evans, A., Schied, C., Keller, A.: Instant neural graphics primitives with a multiresolution hash encoding. ACM Trans. Graph.  \textbf{41}(4),  102:1--102:15 (Jul 2022). \doi{10.1145/3528223.3530127}, \url{https://doi.org/10.1145/3528223.3530127}

\bibitem{article}
Psychogios, K., Helen, L., Melissari, F., Bourou, S., Anastasakis, Z., Zahariadis, T.: Samstyler: Enhancing visual creativity with neural style transfer and segment anything model (sam). IEEE Access  \textbf{PP}, ~1--1 (01 2023). \doi{10.1109/ACCESS.2023.3315235}

\bibitem{yu_and_fridovichkeil2021plenoxels}
{Sara Fridovich-Keil and Alex Yu}, Tancik, M., Chen, Q., Recht, B., Kanazawa, A.: Plenoxels: Radiance fields without neural networks. In: CVPR (2022)

\bibitem{gsinstyle}
Saroha, A., Gladkova, M., Curreli, C., Yenamandra, T., Cremers, D.: Gaussian splatting in style. arXiv preprint arXiv:2403.08498  (2024)

\bibitem{simonyan2015deepconvolutionalnetworkslargescale}
Simonyan, K., Zisserman, A.: Very deep convolutional networks for large-scale image recognition (2015), \url{https://arxiv.org/abs/1409.1556}

\bibitem{igs2gs}
Vachha, C., Haque, A.: Instruct-gs2gs: Editing 3d gaussian splats with instructions (2024), \url{https://instruct-gs2gs.github.io/}

\bibitem{9711417}
Wu, X., Hu, Z., Sheng, L., Xu, D.: Styleformer: Real-time arbitrary style transfer via parametric style composition. In: 2021 IEEE/CVF International Conference on Computer Vision (ICCV). pp. 14598--14607 (2021). \doi{10.1109/ICCV48922.2021.01435}

\bibitem{ye2023gaussian}
Ye, M., Danelljan, M., Yu, F., Ke, L.: Gaussian grouping: Segment and edit anything in 3d scenes (2023)

\bibitem{Yifan_2019}
Yifan, W., Serena, F., Wu, S., Öztireli, C., Sorkine-Hornung, O.: Differentiable surface splatting for point-based geometry processing. ACM Transactions on Graphics  \textbf{38}(6),  1–14 (Nov 2019). \doi{10.1145/3355089.3356513}, \url{http://dx.doi.org/10.1145/3355089.3356513}

\bibitem{coarf}
Zhang, D., Fernandez-Labrador, C., Schroers, C.: Coarf: Controllable 3d artistic style transfer for radiance fields (2024), \url{https://arxiv.org/abs/2404.14967}

\bibitem{stylizedgs}
Zhang, D., Chen, Z., Yuan, Y.J., Zhang, F.L., He, Z., Shan, S., Gao, L.: Stylizedgs: Controllable stylization for 3d gaussian splatting. arXiv preprint arXiv:2404.05220  (2024)

\bibitem{zhang2022arf}
Zhang, K., Kolkin, N., Bi, S., Luan, F., Xu, Z., Shechtman, E., Snavely, N.: Arf: Artistic radiance fields (2022)

\bibitem{feature3dgs}
Zhou, S., Chang, H., Jiang, S., Fan, Z., Zhu, Z., Xu, D., Chari, P., You, S., Wang, Z., Kadambi, A.: Feature 3dgs: Supercharging 3d gaussian splatting to enable distilled feature fields. In: Proceedings of the IEEE/CVF Conference on Computer Vision and Pattern Recognition (CVPR). pp. 21676--21685 (June 2024)

\bibitem{tipeditor}
Zhuang, J., Kang, D., Cao, Y.P., Li, G., Lin, L., Shan, Y.: Tip-editor: An accurate 3d editor following both text-prompts and image-prompts (2024), \url{https://arxiv.org/abs/2401.14828}

\bibitem{surfacesplat}
Zwicker, M., Pfister, H., Baar, J., Gross, M.: Surface splatting. Proceedings of the ACM SIGGRAPH Conference on Computer Graphics  \textbf{2001} (08 2001). \doi{10.1145/383259.383300}

\end{thebibliography}
\end{document}